\pdfoutput=1

\documentclass[11pt]{article}

\usepackage[preprint]{acl}

\usepackage{times}
\usepackage{latexsym}

\usepackage[T1]{fontenc}

\usepackage[utf8]{inputenc}

\usepackage{microtype}

\usepackage{inconsolata}

\usepackage{graphicx}

\usepackage{microtype}
\usepackage{hyperref}
\usepackage{url}
\usepackage{booktabs}
\usepackage{amssymb}
\usepackage{amsthm}
\usepackage{enumitem}
\usepackage{listings}
\usepackage{inconsolata}

\definecolor{codegreen}{rgb}{0,0.6,0}
\definecolor{codegray}{rgb}{0.5,0.5,0.5}
\definecolor{codepurple}{rgb}{0.58,0,0.82}
\definecolor{backcolour}{rgb}{0.95,0.95,0.95}

\setlist[itemize]{leftmargin=*}
\setlist[enumerate]{leftmargin=*}

\lstdefinestyle{mystyle}{
  backgroundcolor=\color{backcolour}, commentstyle=\color{codegreen},
  keywordstyle=\color{magenta},
  numberstyle=\tiny\color{codegray},
  stringstyle=\color{codepurple},
  basicstyle=\ttfamily\footnotesize,
  breakatwhitespace=true,         
  breaklines=true,       
  breakindent=0pt,
  captionpos=b,                    
  keepspaces=true,                 
  numbers=none,                    
  numbersep=5pt,                  
  showspaces=false,                
  showstringspaces=false,
  showtabs=false,                  
  tabsize=2,
  extendedchars=true, 
  literate=%
  {Ö}{{\"O}}1
  {Ä}{{\"A}}1
  {Å}{{\AA{}}}1
  {Ü}{{\"U}}1
  {ß}{{\ss}}1
  {ü}{{\"u}}1
  {ö}{{\"o}}1
  {ä}{{\"a}}1
  {å}{{\aa{}}}1
  {á}{{\'a}}1
  {ã}{{\~a}}1
  {é}{{\'e}}1,
}

\usepackage[strict]{changepage}
\usepackage{xcolor}
\usepackage{framed}
\definecolor{demonstrationshade}{rgb}{0.95,0.95,1}
\definecolor{promptshade}{rgb}{0.95,0.95,1}

\usepackage{tikz}
\setlist[itemize]{leftmargin=*}
\usepackage{makecell}
\usepackage{subfigure}

\usepackage{natbib}
\usepackage{multicol}
\usepackage{multirow}
\usepackage{bm}
\usepackage{xr}
\usepackage{amsmath}
\usepackage[capitalize,noabbrev]{cleveref}
\usepackage[most]{tcolorbox}
\definecolor{block-gray}{gray}{0.85}
\newtcolorbox{myquote}{colback=block-gray,grow to right by=0mm,grow to left by=0mm,
boxrule=0pt,boxsep=0pt,breakable}

\title{Formal-LLM: Integrating Formal Language and Natural Language for Controllable LLM-based Agents}

\author{
 \textbf{Zelong Li},
 \textbf{Wenyue Hua},
 \textbf{Hao Wang},
 \textbf{He Zhu},
 \textbf{Yongfeng Zhang}\\
 Department of Computer Science, Rutgers University, New Brunswick
\\
   \{zelong.li, wenyue.hua, hw488, hz375, yongfeng.zhang\}@rutgers.edu}

\begin{document}
\maketitle
\begin{abstract}
Recent advancements on Large Language Models (LLMs) enable AI Agents to automatically generate and execute multi-step plans to solve complex tasks.
However, since LLM's content generation process is hardly controllable, current LLM-based agents frequently generate invalid or non-executable plans, which jeopardizes the performance of the generated plans and corrupts users' trust in LLM-based agents.
In response, this paper proposes a novel ``Formal-LLM'' framework for LLM-based agents by integrating the expressiveness of natural language and the precision of formal language. Specifically, the framework allows agent developers to express their requirements or constraints for the planning process as an automaton. A stack-based LLM plan generation process is then conducted under the supervision of the automaton to ensure that the generated plan satisfies the constraints, making the planning process controllable. We conduct experiments on both benchmark tasks and practical real-life tasks, and our framework achieves over 50\% overall performance increase, which validates the feasibility and effectiveness of employing Formal-LLM to guide the plan generation of agents, preventing the agents from generating invalid and unsuccessful plans. Further, more controllable LLM-based agents can facilitate the broader utilization of LLM in application scenarios where high validity of planning is essential. The source code of this work is available at \url{https://github.com/agiresearch/Formal-LLM}.
\end{abstract}

\section{Introduction}

Numerous applications have emerged with the rapid development of Large Language Models (LLM). One notable application is the LLM-based agent, which is capable of automatically generating and executing multi-step plans to solve complex tasks.
While the LLM-based agent exhibits creativity, there is a concern about the potential generation of unreasonable and invalid plans, undermining the effectiveness of agents. For example, generating a plan that attempts to process image data using a tool designed for text can lead to errors. Recent studies have pointed out the challenges of LLM-based agents in developing non-executable plans without sufficient human oversight \citep{openagi, yuan2024easytool}. 
Addressing these challenges is crucial for improving agent performance, increasing the generation of valid plans, and maintaining user trust. Various attempts have been made to control LLM text generation, such as incorporating hard constraints \citep{takase2019positional, carlsson2022fine}, soft constraints \citep{gu-etal-2022-distributional, lu2022quark}, or a combination of both \citep{chen2024benchmarking}. However, the focus of controlling LLM-based agents emphasizes the validity of plans and tool use over purely text generation. Some studies \citep{openagi, yuan2024easytool} use LLM as a parser to extract a chain of tools from the generated texts based on prompts, yet satisfactory ratios for valid plans remain elusive.

\begin{figure*}[t]
  \centering
  \includegraphics[width=1\textwidth]{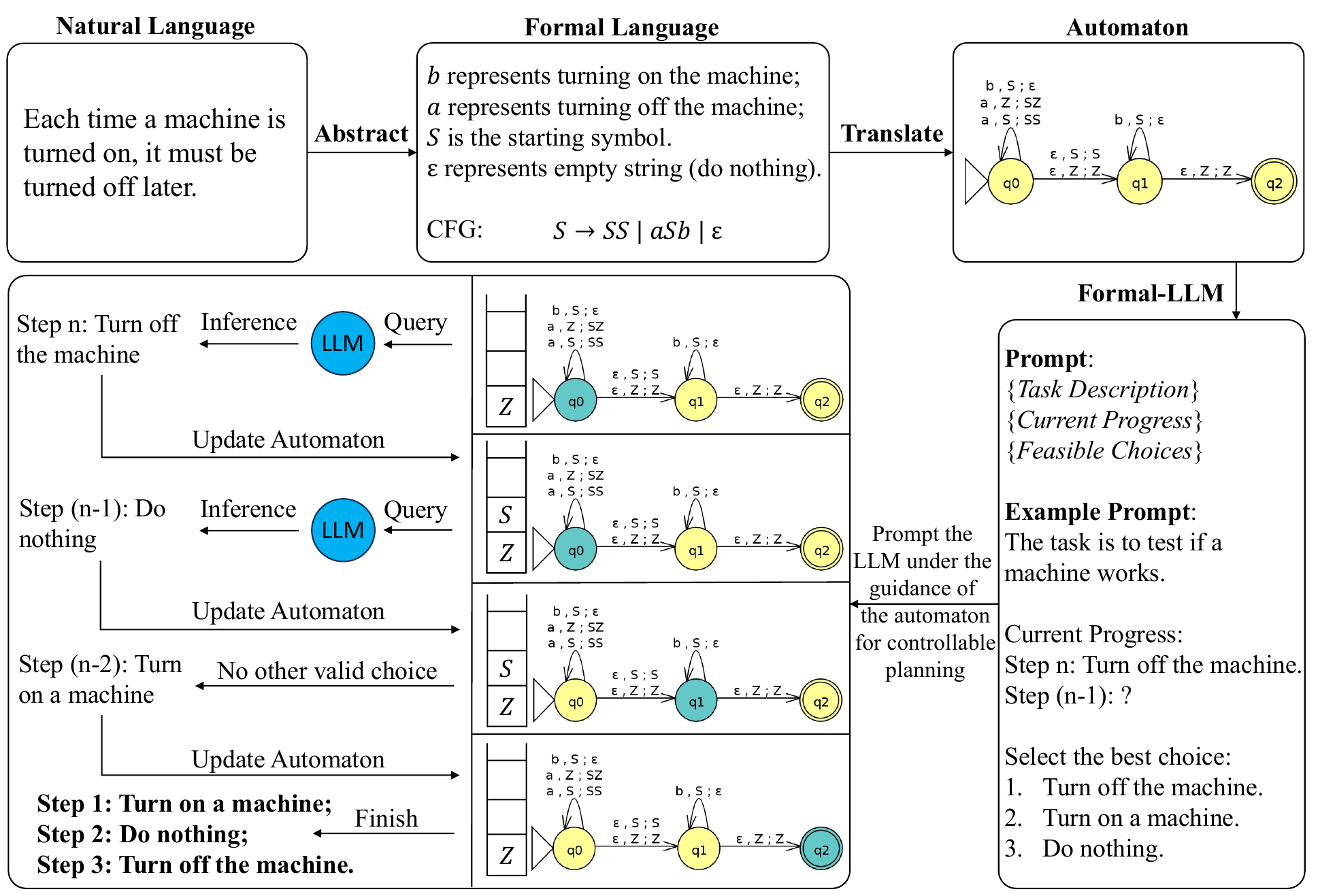}
  \caption{The Formal-LLM workflow with a toy example. To control the LLM-based agent's plan generation, agent developers construct a formal language (context-free grammar, CFG) to represent the natural language constraints. The formal language is then translated into a pushdown automaton (PDA). When LLM conducts planning, it needs to follow the state transition defined by the automaton, which helps to guarantee that the constraint is satisfied in the final generated plan. We choose PDA because some plans exhibit tree-structure (as shown by other examples in the paper), which requires a PDA to generate.
  }
  \label{fig:workflow}
  \vspace{-10pt}
\end{figure*}

To tackle the problems of invalid plan generation, we propose a framework named ``Formal-LLM'', which integrates the expressiveness of natural language and the precision of formal language, as shown by the toy example in Figure \ref{fig:workflow}. 
Specifically, to control the LLM-based agent's plan generation, agent developers construct a context-free grammar (CFG) as the formal language to represent the constraints for the agent. The CFG is then automatically translated into a pushdown automaton (PDA). When LLM conducts planning, it is prompted to follow the state transition defined by the automaton. This is realized by limiting the LLM-based agent's choices at each step to the valid actions defined by the PDA at its current state, which helps to guarantee that the constraint is satisfied in the final generated plan. We choose PDA in this work because some tasks need to be solved by tree-structured plans rather than chain-structured plans, 
and generating tree-structured plans requires a PDA.


Furthermore, we introduce the backtracking mechanism in the Formal-LLM to increase the probability of finding valid plans for the agent, which enables the planning process to return to the previous step when a dead end is reached on the automaton. Besides, traditional LLM-based agent fine-tuning techniques such as Reinforcement Learning from Task Feedback (RLTF) \citep{openagi} rely on the rewards from the agent's plan execution to fine-tune the LLM parameters. However, since many invalid plans may be generated, many rewards are actually not informative for LLM-based agent fine-tuning. Our Formal-LLM approach guarantees that invalid plans are excluded during the agent's plan generation. As a result, our approach helps to increase the amount of valid rewards for LLM fine-tuning, which improves the performance of fine-tuned LLM-based agents.




In our experiments, we implement the Formal-LLM framework across various LLMs, including both open-source and closed-source LLMs. We also test our framework on both benchmark tasks and real-life practical tasks. Specifically, the benchmark tasks involve the LLM-based agent utilizing different tools to solve complex problems through multiple steps. The real-life scenarios encompass daily routines, cooking instructions, 
and commercial risk management, each imposing specific domain-knowledge or common-sense constraints. Our findings demonstrate the Formal-LLM's capability to generate reasonable plans. The framework substantially enhances the overall performance on the benchmark tasks by over 50\% and can always generate executable plans. In real-life scenarios, we provide qualitative analyses into the improvements brought by the Formal-LLM framework, affirming its feasibility and effectiveness in rendering the agents more controllable.



\vspace{-5pt}
\section{Related Work}
\label{sec:related_work}

\subsection{LLM-based AI Agent}

An AI agent is an autonomous entity capable of making decisions and executing actions within a specific environment to effectively tackle diverse complex tasks using AI techniques \citep{ge2023llm, mei2024aios, wang2023survey, xi2023rise}. The emergence of Large Language Models (LLMs), exemplified by the GPT series \citep{radford2019language, brown2020language, openai2023gpt4} and the LLaMA series \citep{touvron2023llama, touvron2023llama2}, has spurred the exploration of LLM-based agents \citep{openagi,huang2022language}. These agents utilize LLMs as their central cognitive component or controller, broadening their perceptual and action capabilities through approaches like multimodal perception and tool usage \citep{schick2023toolformer, openagi, qin2023tool}. In contrast to pre-LLM era AI agents, LLM-based agents showcase creativity, manifesting in the ability to generate innovative ideas without additional learning \citep{franceschelli2023creativity}, indicating a degree of self-directed exploration and decision-making  \citep{xi2023rise}. Despite LLM-based agent's application in diverse real-world scenarios such as software development \citep{li2023camel, qian2023communicative}, scientific research \citep{boiko2023emergent}, and system management \citep{liu2023agentbench}, recent studies have highlighted issues of non-executable plans being generated without adequate human oversight \citep{openagi, yuan2024easytool}. If these plans lack executability, the utility of LLM-based agents in fields that require high validity is compromised, and their unreliability erodes user trust. To address this challenge, we propose integrating natural language and precise automaton during the LLM-based agent's planning process.


\subsection{Controllable LLM Generation}

To the best of our knowledge, there are few works on controllable LLM-based agents, and discussions primarily revolve around controllable LLM text generation. Controlling LLM text generation typically falls into two categories: hard constraints and soft constraints \citep{qin2022cold}. Hard constraints involve constraints on the desired text length, designated keywords for the generated text \citep{takase2019positional, carlsson2022fine}, and constraint vocabulary space in decoding time \citep{hemmer2023lazy, geng2024sketch, geng2023grammar}, while soft constraints restrict the output based on specific semantics, such as sentiments or topics \citep{gu-etal-2022-distributional, lu2022quark, li2020towards}. A recent work \citep{chen2024benchmarking} attempts to integrate both constraints into a unified approach. However, text generation control has disadvantages and problems. First, text generation control could hardly be directly applied to closed-source LLMs due to their need to access the decoding probability \citep{geng2024sketch}. Second, controllable LLM-based agents prioritize the validity of plan and tool use over purely text generation. Recent studies on LLM-based agents have noted limited effectiveness when employing text constraint generation on LLM-based agents and have explored using LLMs as a parser to extract planning information from generated texts \citep{openagi, yuan2024easytool}; but relying heavily on the parser LLM's effectiveness may not yield a satisfactory plan valid rate
. We propose an automaton-guided agent planning to ensure that the generated plans adhere 100\% to the constraints.

\vspace{-5pt}
\section{Preliminary Knowledge}
\label{sec:preliminary}

\subsection{Context-free Grammar}
\label{sec:formal}





Many LLM agents operate tools that take multiple inputs and produce one output. For example, the tool ``Visual Question Answering'' receives an image and a text as the input and outputs a text. The plan involving such tools forms a tree structure, 
and such tree-structured plans require \textbf{Context-Free Grammar (CFG)} to express the rules.

The CFG consists of four components: \textit{terminals} (symbols unable to be replaced), \textit{nonterminals} (symbols subject to replacement), \textit{start symbol} (a unique nonterminal, usually denoted as $S$), and \textit{productions} (rules governing symbol substitution). The CFG production format follows:
\begin{equation}
    A \rightarrow \alpha
  \label{Eq:production}
\end{equation}
where $A$ is a single \textit{nonterminal}, and $\alpha$ is a string comprising any combination of \textit{terminals} and \textit{nonterminals}. $A$ can be replaced by $\alpha$ in any situation.
\begin{equation}
\begin{split}
    & S\rightarrow \varepsilon| aSb
\end{split}
\label{Eq:cfg}
\end{equation}
Eq.\eqref{Eq:cfg} is a CFG example with two productions separated by ``$|$'', where $a, b$ are terminals, $S$ is the start symbol and the only nonterminal, and $\varepsilon$ stands for the empty string. From the start symbol, we can use these two productions to generate infinite strings, such as the empty string, $ab$, and $aabb$. Take the derivation of $aabb$ as an example. We first apply the second production twice ($S\rightarrow aSb \rightarrow aaSbb$) and then the first production once ($aaSbb\rightarrow aabb)$. The formal language, \textbf{Context-Free Language (CFL)}, contains the set of strings of terminals, which can be derived from $S$ in a certain amount of CFG production steps.



\begin{figure}[t]
  \centering
  \includegraphics[width=0.43\textwidth]{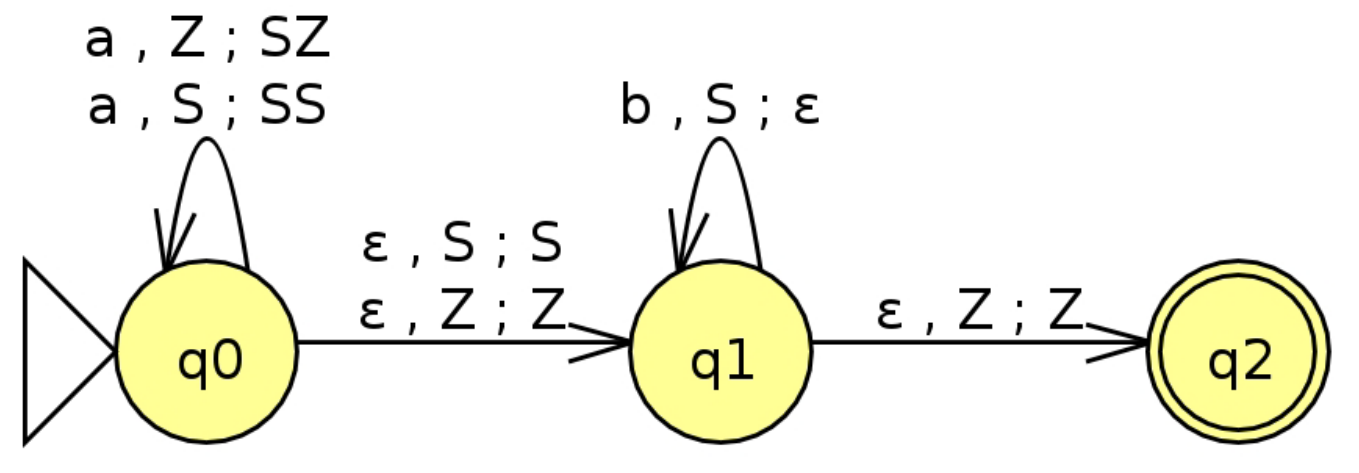}
  \vspace{-10pt}
  \caption{A Pushdown Automaton (PDA) example. 
  }
  \label{fig:pda}
  \vspace{-10pt}
\end{figure}

\begin{figure}[t]
  \centering
  \includegraphics[width=0.48\textwidth]{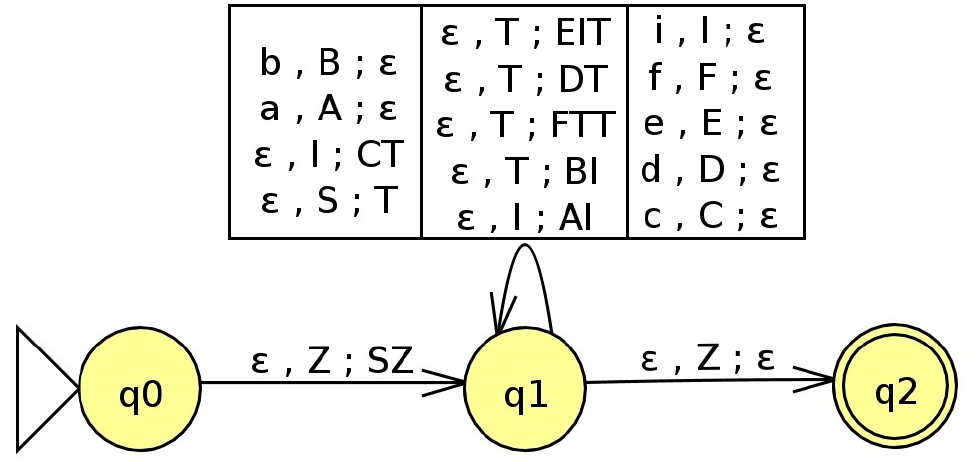}
  \vspace{-20pt}
  \caption{An equivalent PDA to the CFG of the combination of Eq.\eqref{Eq:agi_cfg} and Eq.\eqref{Eq:agi_cfg_second}. The lowercase letters $a$ to $f$ are $a_i$ to $f_j$ in Eq.\eqref{Eq:agi_cfg}. For example, the transit $(b, B; \varepsilon)$ implies the PDA removes the stack top $B$ and remains in state $q_1$ if the PDA is at state $q_1$, the stack top is $B$, and the next string letter is $b_1/b_2/b_3$.}
  \label{fig:agi_cfg}
\vspace{-20pt}
\end{figure}


\subsection{Pushdown Automaton}
\label{sec:automaton}

Every CFG can be transformed into an equivalent nondeterministic \textbf{Pushdown Automaton (PDA)} with an algorithm \citep{hopcroft2001introduction}. A PDA is a machine that moves through a series of states based on a given string input, a transition function, and the top element of a stack. The automaton reads the letters in the input sequentially, and if there exists a path leading the machine to an accept state right after consuming the entire string, the input string is considered an \textit{accepting word}. The set of all accepting words is referred to the CFL accepted by the automaton. Since the PDA process is more intuitive and automated than CFG for guiding the prompting of LLMs, our framework is designed on top of PDA. 

Figure \ref{fig:pda} shows how PDA works as an example. This automaton includes a stack alphabet ($\{S, Z\}$) with an initial stack symbol $Z$, a start state $q_0$, an accept state set $\{q_2\}$, and a transit function over states. The transit functions are in the form of triplets. For example, the edge $(a, Z; SZ)$ from $q_0$ to $q_0$ means that if the stack top is $Z$ and the next letter of the string is $a$, the PDA remains in $q_0$ state and the stack top changes from $Z$ to $SZ$ ($S$ is the stack top now).

This PDA is equivalent to the CFG in Eq.\eqref{Eq:cfg}, i.e., every accepting word of this PDA belongs to the CFL defined in Eq.\eqref{Eq:cfg}. We still use $aabb$ as an example to show the PDA mechanism. Initially, the PDA is at $q_0$ and the stack only contains $Z$. We first apply $(a, Z; SZ)$ when receiving the first letter $a$, and the PDA remains in $q_0$ with the stack as $SZ$. Then, when consuming the next letter $a$, we use $(a, S; SS)$ and $(\varepsilon, S; S)$, move to $q_1$ state, and change the stack to $SSZ$. Next, the PDA remains in $q_1$ and applies $(b, S; \varepsilon)$ function twice to consume the two $b$'s in the input string, and at the same time, the stack transforms from $SSZ$ to $Z$. Finally, the automaton use $(\varepsilon, Z; Z)$ to move to the accept state $q_2$. Since the machine finishes reading the input string, $aabb$ is an accepting word of this CFL. $aabbb$ is not an accepting word because, after reading $aabb$, the automaton is at state $q_1$ with the stack as $Z$, and there is no valid function to consume the final $b$ with $Z$ as the stack top.

\section{The Formal-LLM Framework}
\label{sec:framework}

\subsection{Motivation and Challenge}
\label{sec:motivation}

Natural language is easily understandable by humans, but it may also lack precision in certain application scenarios. In contrast, formal language is defined in clear mathematical and machine readable forms, which could be hard to comprehend for humans, but it also possesses great precision. Thus, we aim to effectively integrate the benefits of both natural and formal languages through the pushdown automaton (PDA) for more controllable planning of LLM-based agents in scenarios requiring precise and valid plans. 


Constructing our Formal-LLM framework presents non-trivial challenges.
First, LLMs may struggle to directly comprehend and process formal language due to the limited CFG- or PDA-related corpus in pre-training data. Hence, natural language prompts are necessary to effectively describe the status of the automaton, bridging the gap between formal and natural language for LLM's planning. Given that the automaton has the potential to generate an infinite number of plans, the second challenge arises because the automaton only ensures the validity and executability of the plans by adhering to defined constraints, but cannot guarantee the performance and optimality of the generated plan. Therefore, leveraging the solid natural language understanding ability of LLM and other methods, such as fine-tuning and backtracking mechanisms, becomes crucial for enhancing planning performance.

\subsection{Formulating Constraint to Automaton}
\label{sec:formulating}

The overall flow of our framework is presented in Figure \ref{fig:workflow}. We begin by illustrating the process of converting constraints into formal language and automata for benchmark tasks, using the OpenAGI benchmark \citep{openagi} as an example, which incorporates a set of domain expert models as tools and lists a series of intricate problems that cannot be addressed with a single tool. These tools are categorized into six primary groups according to the input and output modalities, as indicated in Table \ref{Table:openagi_tool} in the Appendix.

If the plans for the benchmark tasks generated by the LLM-based agent deviate from the expected data format, the execution performance degrades significantly or even results in errors. For instance, the \textit{Colorization} tool operates on image data as input and output. Since image data is represented as a 3-D tensor, providing a 1-D tensor text string as input could lead to errors, due to incorrect data dimensions. Thus, the entire plan becomes invalid and cannot be executed. In such cases, natural language may struggle to articulate these constraints precisely and concisely, but formal language can express them straightforwardly.

The CFG in Eq.\eqref{Eq:agi_cfg} outlines the constraints on data formats with Polish prefix notation: $T$ for text and $I$ for image, capital letters $A$ to $F$ for the six types of tools, and lowercase letters such as $a_i\in A$ or $f_j\in F$ for specific tools in the order of Table \ref{Table:openagi_tool}.
\begin{equation}
\begin{alignedat}{7}
& I \rightarrow AI | CT \qquad && T \rightarrow BI | DT | EIT | FTT \\
& A \rightarrow a_1 | a_2 | a_3 | a_4
 \qquad && B \rightarrow b_1 | b_2 | b_3\\
& C \rightarrow c_1 \qquad && D \rightarrow d_1 | d_2 | d_3 | d_4 | d_5
  \\ & E \rightarrow e_1 \qquad && F \rightarrow f_1
\end{alignedat}
\label{Eq:agi_cfg}
\end{equation}

where $I \rightarrow AI | CT$ means $I$ can be replaced by $AI$ or $CT$, because $A$ is an image-to-image tool and $C$ is a text-to-image tool. As a result, when $A$ is applied on an image ($AI$) or $C$ is applied on a text ($CT$), the result is still an image.
Considering a specific task, such as ``Given blurry gray-scale images, how to return the object names in English step by step?'', we know that the task's inputs are images and the final outputs should be texts. To formalize this, we add two constraints to Eq.\eqref{Eq:agi_cfg}:
\begin{equation}
\begin{alignedat}{0}
    & S \rightarrow T \qquad && I \rightarrow i
\end{alignedat}
\label{Eq:agi_cfg_second}
\end{equation}
i.e., the CFG starts with the symbol $T$, implying the final text format, and uses the lowercase letter $i$ to represent the input image. The combination of Eq.\eqref{Eq:agi_cfg} and Eq.\eqref{Eq:agi_cfg_second} renders the data format constraints of the specific task. For instance, $e_1a_1ib_1i$ is an accepted word, representing a valid plan: ``Utilize \textit{Image Classification} ($b_1$) on the input image to obtain text and apply \textit{Colorization} ($a_1$) on the input image to generate the output image. Then, employ \textit{Visual Question Answering} ($e_1$) with the text and output image to derive the final text.'' As mentioned in Section \ref{sec:automaton}, the CFG can be equivalently converted into a PDA in Figure \ref{fig:agi_cfg}. These steps allow us to transform the constraints into automata for subsequent utilization.

\begin{figure*}[t]
  \centering
  \includegraphics[width=0.8\textwidth]{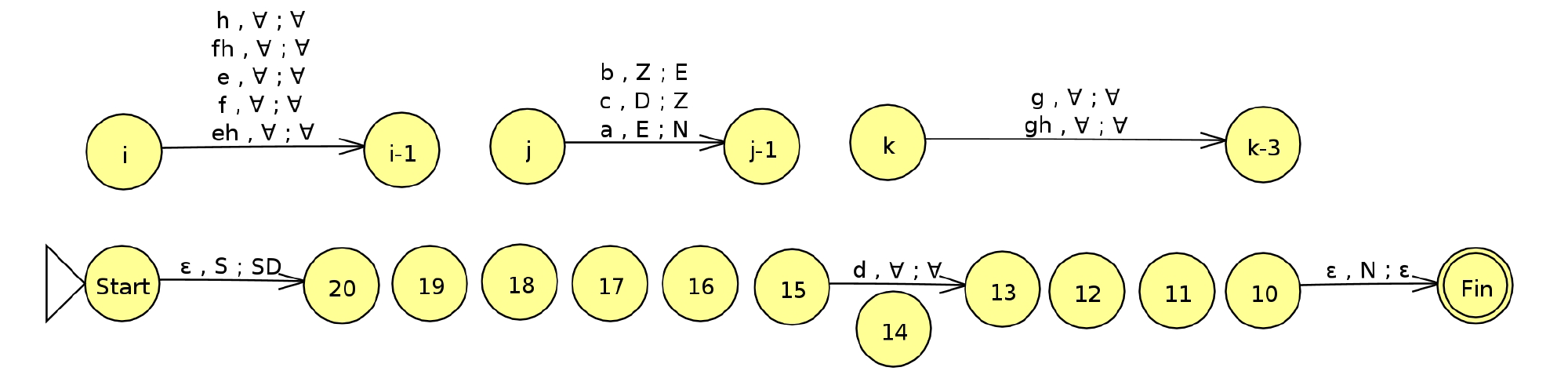}
  \caption[caption for daily]{The PDA includes 13 states, $\{10, 11, ..., 19, 20, Start, Fin\}$, with $Start$ as the start state and $\{Fin\}$ as the accept state set. The lowercase letter $a$ is for breakfast, $b$ is for lunch, $c$ is for supper, $d$ is for playing basketball, $e$ is for grocery shopping, $f$ is for house cleaning, $g$ is for homework, and $h$ is for laundry. The stack symbols $D, Z, E, N$ represent 3, 2, 1, and 0 meals remaining to plan, respectively. $\forall$ represents any stack symbol. The variable $i$ is enumerated from 20 to 11, $j$ is enumerated from 20 to 11 except 13, and $k$ is enumerated from 20 to 13.}
  \label{fig:daily}
  \vspace{-10pt}
\end{figure*}

In certain scenarios, creating an automaton can be more straightforward than formulating a formal language. Section \ref{sec:daily} describes a real-life task for daily work planning. The constraints in this case revolve around time, allowing us to represent time as distinct states, as depicted in Figure \ref{fig:daily}. The automaton initiates the design of an activity that concludes at 20:00, and a valid plan is generated when the automaton reaches 10:00 while being in a state where three meals have been consumed. In this case, we design a PDA without a CFG.


\subsection{Formal-LLM Prompts and Planning from Automaton}
\label{sec:guidance}

In Section \ref{sec:formulating}, we convert the natural language planning constraints into a PDA, where any accepted word represents a valid and executable plan. The reason is that LLM could struggle to directly comprehend or process the CFG due to limited exposure to formal language during pre-training. Hence, we employ natural language prompts for LLM to comprehend the task and generate plans that can be easily read by humans; simultaneously, we use the PDA to guide the process of generating natural language plans. 

The automaton initiates the plan generation process from the \textit{start state}. In situations where multiple viable options are available for the automaton to proceed, a prompt is created to inquire the LLM, as shown in Figure \ref{fig:workflow}. This prompt encompasses a comprehensive \{task description\} to mitigate potential forgetting issues of the LLM \citep{hua2023war}, details about the \{current progress\}, and \{feasible choices\} determined by the PDA. For example, when the PDA in Figure \ref{fig:agi_cfg} is in state $q_1$ with the symbol $I$ at the top of the stack, three viable transits: $(\varepsilon, I; AI)$, $(\varepsilon, I; CT)$, and $(i, I; \varepsilon)$ are included in the \{feasible choices\} of the prompt. A prompt example is provided in Appendix \ref{sec:agi_prompt}.

\begin{figure}[t]
  \centering
  \includegraphics[width=0.35\textwidth]{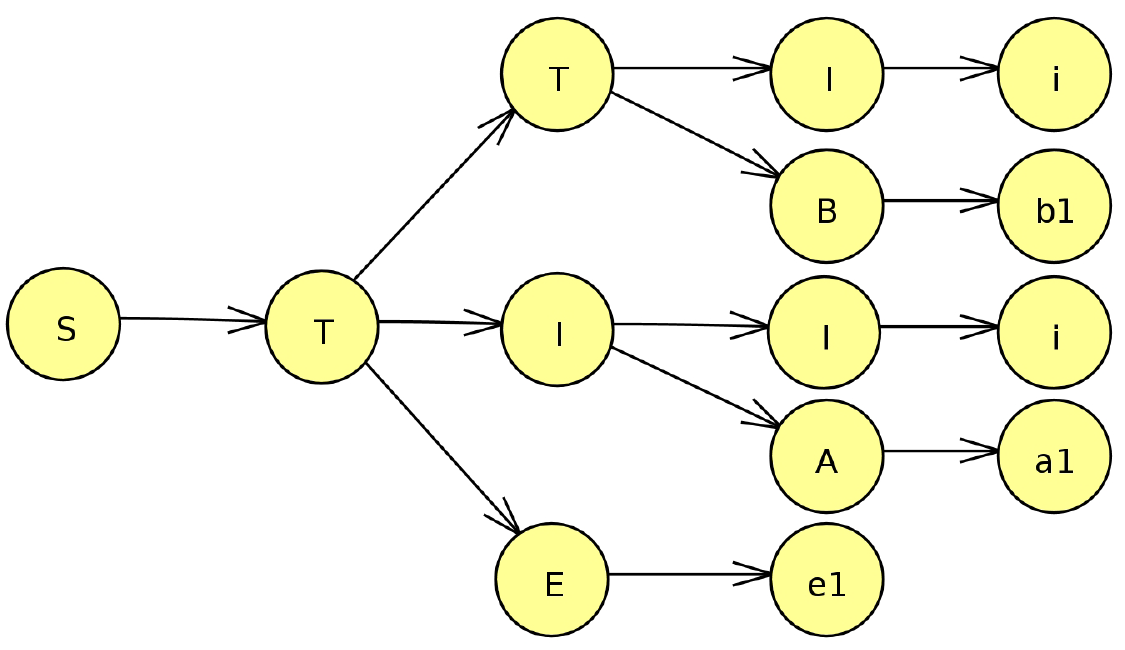}
  \vspace{-5pt}
  \caption[caption for syntax tree]{The derivation tree for $e_1a_1ib_1i$ for the example benchmark task in Section \ref{sec:formulating}.}
  \label{fig:syntax}
  \vspace{-10pt}
\end{figure}

Another advantage of employing PDA for LLM-based agent planning is its ability to articulate plans that involve tools with many-input-single-output (e.g., Image-Text pair as input, Text as output), assuming the number of inputs of each tool is deterministic. To illustrate, we depict the derivation tree of the string $e_1a_1ib_1i$ in Figure \ref{fig:syntax}. Treating the tools $a_1$ and $b_1$ as functions, the string is functionally equivalent to $t = e_1 \big(a_1(i), b_1(i)\big)$. When the PDA finishes the substitution of symbols $E$ and $I$, it can well know that the stack symbol $T$ is derived from $(\varepsilon, T; EIT)$ and matches the input of tool $e_1$, because they are pushed into the stack at the same time. Finally, we can identify the tools that will operate on this image/text data in the subsequent step and incorporate this information in the prompt, enabling the LLM to make informed decisions.


\subsection{Reinforcement Learning from Task Feedback and Backtracking Mechanism}
\label{sec:mechanisum}

An accepting word of an automaton can be infinitely long. However, we anticipate that the plan proposed by the LLM-based agent can be executed within a finite number of steps. Consider the PDA in Figure \ref{fig:pda}, any length of the string in the form of $a^nb^n$ where $n\geq 0$ is an accepting word. Similarly, in planning tasks, tools can be employed multiple times, such as applying the \textit{Text Summarization} tool to the original text and infinitely repeating the same tool on the output. However, such planning behavior may be nonsensical and inefficient. Hence, for the tasks outlined in this paper, each tool is restricted to a single use for a given task, preventing infinite planning for both benchmark and real-life scenarios.

Upon imposing constraints on the tool usage, a new challenge emerges: the automaton may encounter a dead-end during plan generation. Consider the formal language in Eq.\eqref{Eq:agi_cfg} and Eq.\eqref{Eq:agi_cfg_second}. Upon analyzing the language, we observe that the only means of transitioning from the symbol $T$ to $I$ (and subsequently to a terminal $i$) involve $B$-type (Image-in, Text-out) and $E$-type (Image-Text-pair-in, Text-out) tools. If these two tool types are exhausted, yet there are still $T$ symbols in the stack, the automaton will inevitably fail to reach the accept state $q_2$, as no methods remain to clear the symbol $T$. Finally, a dead-end situation may arise following the imposition of tool usage limits.

To address this challenge, we propose to document the automaton's details at each step, encompassing the current state and stack, and the part of the word that has been generated. When the automaton confronts a dead-end
, we initiate backtracking to the preceding steps until there is an unexplored branch. The backtracking involves retracing the automaton to the prior step, reinstating the automaton details based on the recorded stack, and eliminating the dead-end branch from the prior choice list. Thus, the automaton is capable and guaranteed to generate a valid plan if it exists. 

The design above only guarantees the validity of the generated plans
, but not the optimal plan. Thus, as an additional enhancement in plan quality, particularly when employing open-source LLMs, we integrate Reinforcement Learning from Task Feedback (RLTF) \citep{openagi} following the application of our Formal-LLM framework. After the LLM-based agent generates a plan for a training task, the plan is executed on benchmark data to assess its performance. Then, the performance is used as a reward for RL to update the parameters of the LLM. Our framework ensures that invalid plans are always excluded during the LLM-based agent's plan generation. Therefore, the number of valid rewards for LLM fine-tuning increases, and our approach can improve the performance of fine-tuned LLM-based agents. 

\section{Experiments}
\label{sec:experiment}

\subsection{Backbone Large Language Model (LLM)}

We test our Formal-LLM framework on these 
closed-source LLMs:
\begin{itemize}[itemsep=0pt, topsep=0pt]
    \item \textbf{GPT-3.5-turbo} \citep{brown2020language} is a generative pre-trained transformer of OpenAI.
    \item \textbf{Claude-2} \citep{claude2} is a transformer LLM of Anthropic.
    \item \textbf{GPT-4} \citep{openai2023gpt4} is a follow-up version of GPT-3.5.
\end{itemize}

and these open-source LLMs:
\begin{itemize}[itemsep=0pt, topsep=0pt]
    \item \textbf{Flan-T5-Large} \citep{chung2022scaling} is a language model with 770 million parameters.
    \item \textbf{Vicuna-7B} \citep{chiang2023vicuna} is a 7-billion size chatbot trained by fine-tuning the LLaMA model \citep{touvron2023llama}
    .
    \item \textbf{LLaMA-2-13B} \citep{touvron2023llama2} is a successor to the 13-billion size LLaMA model.
\end{itemize}

\vspace{-7pt}
\subsection{Learning Schema of LLMs}

We adopt the following LLM learning schema:

\begin{itemize}[itemsep=0pt, topsep=0pt]
    \item \textbf{Zero-shot Learning (Zero)} directly inputs the prompt to the LLM.
    \item \textbf{Few-shot Learning (Few)} presents a set of high-quality examples in the prompt, each consisting of input and desired output on the target task. 
    \item \textbf{Reinforcement Learning from Task Feedback (RLTF)} applies text constraint generation, executes the plan, and takes its performance as the reward to optimize the LLM by RL.
    \item \textbf{Formal-LLM (F-LLM)} is our proposed framework, using automaton to control planning.
    \item \textbf{Formal-LLM plus RLTF (F-LLM+RLTF)} applies our Formal-LLM framework on top of RLTF without text constraint generation. We use automaton to exclude invalid plans.
\end{itemize}

Specifically, we utilize \textbf{Zero, Few}, and \textbf{F-LLM} frameworks for closed-source LLMs, as these frameworks do not need to modify the LLM's parameters. Regarding open-source LLMs, we compare \textbf{RLTF}, \textbf{F-LLM} and \textbf{F-LLM+RLTF}, given that \textbf{RLTF} surpasses \textbf{Zero} and \textbf{Few}, as exhaustively discussed in recent work \citep{openagi}.

\begin{table*}[t]
\small
    \centering
    \setlength{\tabcolsep}{1pt}
    \begin{tabular}{c|ccc|ccc|ccc}
    \hline
      \multirow{2.5}{*}{Metrics / Task} & \multicolumn{3}{c|}{GPT-3.5-turbo} & \multicolumn{3}{c|}{Claude-2} & \multicolumn{3}{c}{GPT-4} \\
      \cline{2-4} \cline{5-7} \cline{8-10} 
      & Zero & Few & F-LLM (Ours) & Zero & Few & F-LLM (Ours) & Zero & Few & F-LLM (Ours)  \\
      \hline
      \% of Valid Plans & 29\% & 71\% & \textbf{100\%} & 29\% & 47\% & \textbf{100\%} & 53\% & 76\% & \textbf{100\%}  \\
      \hline
      Task 1 (CLIP Score) & 0.0 & 0.0 & \textbf{0.3056} & 0.0 & 0.2543 & \textbf{0.3056} & 0.0 & 0.3055 & \textbf{0.3056} \\
      Task 2 (BERT Score) & 0.1914 & 0.3820 & \textbf{0.6364} & 0.2111 & 0.5038 & \textbf{0.6275} & 0.2076 & \textbf{0.6307} & 0.5102 \\
      Task 3 (ViT Score) & 0.2437 & \textbf{0.7497} & 0.6470 & 0.4082 & 0.5416 & \textbf{0.7137} & 0.5058 & 0.6480 & \textbf{0.7689} \\
      Task X & 0.0 & 0.0 & \textbf{0.0658} & 0.0 & 0.0 & \textbf{0.2799} & 0.0 & 0.0 & \textbf{0.2876} \\
      Average over tasks & 0.1443 & 0.3345 & \textbf{0.4846} & 0.1838 & 0.3773 & \textbf{0.5420} & 0.1992 & 0.4662 & \textbf{0.4914} \\
      \hline
    \end{tabular}
    \caption{Benchmark task performances under different settings for three closed-source LLMs. Zero is for Zero-shot Learning, Few is for Few-shot Learning, and F-LLM is for Formal-LLM. The boldface numbers denote the highest score under each task type using the same LLM.}
    \label{Table:closed_source}
    \vspace{-10pt}
\end{table*}

\begin{table*}[t]
\small
    \centering
    \setlength{\tabcolsep}{0.5pt}
    \begin{tabular}{c|ccc|ccc|ccc}
    \hline
      \multirow{2.5}{*}{Metrics / Tasks} & \multicolumn{3}{c|}{Flan-T5-Large} & \multicolumn{3}{c|}{Vicuna-7B} & \multicolumn{3}{c}{LLaMA-2-13B} \\
      \cline{2-4} \cline{5-7} \cline{8-10} 
      & RLTF & F-LLM & F-LLM+RLTF & RLTF & F-LLM & F-LLM+RLTF & RLTF & F-LLM & F-LLM+RLTF \\
      \hline
      \% of Valid Plans & 24\% & \textbf{100\%} & \textbf{100\%} & 29\% & \textbf{100\%} & \textbf{100\%} & 47\% & \textbf{100\%} & \textbf{100\%}  \\
      \hline
      Task 1 (CLIP Score) & 0.0 & \textbf{0.3049} & \textbf{0.3049} & 0.0 & 0.3122 & \textbf{0.3139} & 0.0610 & 0.1601 & \textbf{0.3060} \\
      Task 2 (BERT Score) & 0.3327 & 0.5164 & \textbf{0.5287} & 0.1475 & \textbf{0.4948} & 0.4673 & 0.1611 & 0.4220 & \textbf{0.5565} \\
      Task 3 (ViT Score) & 0.6632 & 0.6264 & \textbf{0.7469} & 0.6958 & 0.5948 & \textbf{0.8618} & \textbf{0.7106} & 0.7043 & 0.6808 \\
      Task X & 0.0 & 0.0728 & \textbf{0.4046} & 0.0 & \textbf{0.4127} & 0.4029 & 0.0 & 0.3846 & \textbf{0.4163} \\
      Average over tasks & 0.3111 & 0.4451 & \textbf{0.5321} & 0.2009 & 0.4824 & \textbf{0.5162} & 0.3101 & 0.4498 & \textbf{0.5390} \\
      \hline
    \end{tabular}
    \caption{Benchmark task performances under different settings for three open-source LLMs. RLTF is for Reinforcement Learning from Task Feedback, F-LLM is for Formal-LLM, and F-LLM+RLTF is for using the generated plans by F-LLM to calculate the reward for RLTF. The boldface numbers denote the highest score under each task type using the same LLM.}
    \label{Table:open_source}
    \vspace{-10pt}
\end{table*}

\subsection{Experimentation Datasets}

\subsubsection{Benchmark Datasets}

We conduct experiments on two benchmark datasets, i.e., \textbf{OpenAGI} \citep{openagi} and \textbf{TravelPlanner} \citep{xie2024travelplanner}.
Due to space limit, we provide the results on OpenAGI in the following and provide the results on TravelPlanner in Appendix Section \ref{sec:travel_planner}.
The OpenAGI benchmark tasks are categorized based on their output type and ground-truth label type (\textbf{Task 1}, \textbf{2}, and \textbf{3}). 
Then, based on different task types, different metrics are employed to gauge the performance: \textbf{CLIP Score} \citep{hessel2021clipscore}, assessing the similarity between text and image, is utilized for Text-to-Image tasks; \textbf{BERT Score} \citep{bert-score}, evaluating text generation with BERT, is applied when both data labels and the expected outputs are texts; and \textbf{ViT Score} \citep{wu2020visual} gauges the similarity between the image label and image output.
Additionally, we construct \textbf{Task X}, which is a subset of ``Task 1 $\cup$ Task 2 $\cup$ Task 3'' that requires a tree-structured plan 
due to the use of many-input-single-ouput tools, such as \textit{Question Answering}. Task X is used to test the complex planning ability of our Formal-LLM framework,
and is evaluated using the corresponding metric.
The tools are detailed in Table \ref{Table:openagi_tool} and example tasks of each category are in Table \ref{Table:task_example} in the Appendix.

\subsubsection{Real-life Practical Tasks}

We also experiment with real-life planning scenarios, encompassing daily plans, cooking recipes, and risk management, where validity and rationality are crucial. 
In these scenarios, the concept of tool is generalized to include various types of steps in a plan, such as events, actions, or activities,
as they can take diverse forms supporting the execution of the plan. 
For example, a tool in the daily planning task may be an activity such as \textit{taking breakfast}. We provide qualitative analyses between the \textbf{Zero} and \textbf{F-LLM} learning schema using GPT-4 backbone for practical tasks. We use GPT-4 for the experiment of practical tasks because other LLMs can hardly generate readable plans for these tasks, and we test under Zero and F-LLM due to the limited sample size and that these two learning frameworks do not require accessing to the LLM parameters.


\subsection{Experimental Analysis}

The experiment results on the benchmark tasks are shown in Table \ref{Table:closed_source} and Table \ref{Table:open_source}, referring to the closed-source and open-source LLMs, respectively. Each row stands for a type of task, each column represents the learning schema of an LLM-based agent, and every three columns are the results of the same LLM. For closed-source LLMs, among the three learning schemata without revising the LLM parameter, almost all the best scores under each type of task belong to our F-LLM framework.
For open-source LLMs, our F-LLM without fine-tuning is already better than RLTF (the best schema in the OpenAGI platform \citep{openagi}) in most cases, 
except for the ViT Score (Task 3), because
the Type-3 tasks take the most significant portion of the RLTF's fine-tuning data, and thus RLTF is adequately optimized on these tasks. The performance gain of our F-LLM framework comes from the 100\% executable plans and supporting tree-structured planning for the difficult tasks in Task X. As a comparison, the best open-source LLM (LLaMA-2-13B with RLTF) can only generate 47\% executable plans, and the best closed-source LLM (GPT-4 with few shot) can generate 76\% executable plans. 
Additionally, none of the baselines can handle the Task X (Score = 0.0). Due to the increased amounts of valid rewards from the 100\% executable plans with our framework, the F-LLM+RLTF approach enables boosted performance, as demonstrated by its better scores in Table \ref{Table:open_source}. 
To summarize, the benchmark experiment results show that our Formal-LLM framework is an effective method to integrate the benefits of both natural and formal languages for more controllable and valid planning of LLM-based agents.

\subsection{Case Study}

We show the complete results of real-life examples by applying our Formal-LLM to the GPT models 
in the Appendix. From the results, after applying our Formal-LLM, the generated plan from the GPT-based agent is more complete, reasonable and specific to the case. In the daily plan example, the agent fails to fit all the activities into the plan without the Formal-LLM, while it can achieve this goal by applying the Formal-LLM. Still take Figure \ref{fig:daily} as an example, our Formal-LLM can limit the planning to 10:00$\sim$20:00, while without Formal-LLM, the agent schedules activities after 20:00, even if we mention the constraint ``\textit{Generate a plan for all the activities between 10:00 and 20:00}'' in natural language, which shows the advantages of formal language guided planning. 
For the cooking recipe example, without strict automaton constraint, GPT may generate steps such as ``\textit{stir-fry the Chinese broccoli until it changes color}'' after ``\textit{once the water boils, add the Chinese broccoli}''. This is unreasonable because 
broccoli cannot be stir-fried in boiling water. For the risk management example, the generated plan without our framework is too general that it could be used for any other two companies. However, the plan after applying Formal-LLM is more specific and focuses on the potential antitrust risk between Microsoft and Blizzard. 
Thus, the Formal-LLM can use the automaton describing the constraints to generate higher-quality plans.

\section{Conclusions and Future Work}
\label{sec:conclusions}

In this study, we introduce the innovative Formal-LLM framework for LLM-based agents, combining LLM's power of natural language comprehension with the precision of formal language. Our experiments, encompassing benchmark tasks and real-life practical scenarios, affirm the feasibility and effectiveness of employing automaton to control the agent's generation of valid plans. More controllable LLM-based agents can augment the potential utilization of LLM in applications where high validity of planning is important.

Our work has several potential extensions. First, automating the translation of natural language into formal language for agents could further improve the framework.
Additionally, this work focuses on LLM plan generation based on formal language. Another important problem to explore is LLM plan verification based on formal language.

\section*{Acknowledgment}
We thank Harun Taha Kepenek for providing helpful suggestions to improve this research paper.

\bibliography{colm2024_conference}

\begin{thebibliography}{52}
\providecommand{\natexlab}[1]{#1}

\bibitem[{Boiko et~al.(2023)Boiko, MacKnight, and Gomes}]{boiko2023emergent}
Daniil~A. Boiko, Robert MacKnight, and Gabe Gomes. 2023.
\newblock \href {https://arxiv.org/abs/2304.05332} {Emergent autonomous scientific research capabilities of large language models}.
\newblock \emph{Preprint}, arXiv:2304.05332.

\bibitem[{Brown et~al.(2020)Brown, Mann, Ryder, Subbiah, Kaplan, Dhariwal, Neelakantan, Shyam, Sastry, Askell et~al.}]{brown2020language}
Tom Brown, Benjamin Mann, Nick Ryder, Melanie Subbiah, Jared~D Kaplan, Prafulla Dhariwal, Arvind Neelakantan, Pranav Shyam, Girish Sastry, Amanda Askell, et~al. 2020.
\newblock Language models are few-shot learners.
\newblock \emph{Advances in neural information processing systems}, 33:1877--1901.

\bibitem[{Carion et~al.(2020)Carion, Massa, Synnaeve, Usunier, Kirillov, and Zagoruyko}]{carion2020end}
Nicolas Carion, Francisco Massa, Gabriel Synnaeve, Nicolas Usunier, Alexander Kirillov, and Sergey Zagoruyko. 2020.
\newblock End-to-end object detection with transformers.
\newblock In \emph{European conference on computer vision}, pages 213--229. Springer.

\bibitem[{Carlsson et~al.(2022)Carlsson, {\"O}hman, Liu, Verlinden, Nivre, and Sahlgren}]{carlsson2022fine}
Fredrik Carlsson, Joey {\"O}hman, Fangyu Liu, Severine Verlinden, Joakim Nivre, and Magnus Sahlgren. 2022.
\newblock Fine-grained controllable text generation using non-residual prompting.
\newblock In \emph{Proceedings of the 60th Annual Meeting of the Association for Computational Linguistics (Volume 1: Long Papers)}, pages 6837--6857.

\bibitem[{Chen et~al.(2024)Chen, Xu, Wang, Liu, and Mao}]{chen2024benchmarking}
Yihan Chen, Benfeng Xu, Quan Wang, Yi~Liu, and Zhendong Mao. 2024.
\newblock \href {https://arxiv.org/abs/2401.00690} {Benchmarking large language models on controllable generation under diversified instructions}.
\newblock \emph{Preprint}, arXiv:2401.00690.

\bibitem[{Chiang et~al.(2023)Chiang, Li, Lin, Sheng, Wu, Zhang, Zheng, Zhuang, Zhuang, Gonzalez et~al.}]{chiang2023vicuna}
Wei-Lin Chiang, Zhuohan Li, Zi~Lin, Ying Sheng, Zhanghao Wu, Hao Zhang, Lianmin Zheng, Siyuan Zhuang, Yonghao Zhuang, Joseph~E Gonzalez, et~al. 2023.
\newblock Vicuna: An open-source chatbot impressing gpt-4 with 90\%* chatgpt quality.

\bibitem[{Chung et~al.(2022)Chung, Hou, Longpre, Zoph, Tay, Fedus, Li, Wang, Dehghani, Brahma et~al.}]{chung2022scaling}
Hyung~Won Chung, Le~Hou, Shayne Longpre, Barret Zoph, Yi~Tay, William Fedus, Yunxuan Li, Xuezhi Wang, Mostafa Dehghani, Siddhartha Brahma, et~al. 2022.
\newblock \href {https://arxiv.org/abs/2210.11416} {Scaling instruction-finetuned language models}.
\newblock \emph{Preprint}, arXiv:2210.11416.

\bibitem[{Claude-2(2023)}]{claude2}
Claude-2. 2023.
\newblock Model card and evaluations for claude models.

\bibitem[{Conde et~al.(2022)Conde, Choi, Burchi, and Timofte}]{conde2022swin2sr}
Marcos~V Conde, Ui-Jin Choi, Maxime Burchi, and Radu Timofte. 2022.
\newblock Swin2sr: Swinv2 transformer for compressed image super-resolution and restoration.
\newblock In \emph{European Conference on Computer Vision}, pages 669--687.

\bibitem[{Dosovitskiy et~al.(2021)Dosovitskiy, Beyer, Kolesnikov, Weissenborn, Zhai, Unterthiner, Dehghani, Minderer, Heigold, Gelly, Uszkoreit, and Houlsby}]{dosovitskiy2020vit}
Alexey Dosovitskiy, Lucas Beyer, Alexander Kolesnikov, Dirk Weissenborn, Xiaohua Zhai, Thomas Unterthiner, Mostafa Dehghani, Matthias Minderer, Georg Heigold, Sylvain Gelly, Jakob Uszkoreit, and Neil Houlsby. 2021.
\newblock An image is worth 16x16 words: Transformers for image recognition at scale.
\newblock \emph{ICLR}.

\bibitem[{Franceschelli and Musolesi(2023)}]{franceschelli2023creativity}
Giorgio Franceschelli and Mirco Musolesi. 2023.
\newblock \href {https://arxiv.org/abs/2304.00008} {On the creativity of large language models}.
\newblock \emph{Preprint}, arXiv:2304.00008.

\bibitem[{Ge et~al.(2023{\natexlab{a}})Ge, Hua, Mei, Ji, Tan, Xu, Li, and Zhang}]{openagi}
Yingqiang Ge, Wenyue Hua, Kai Mei, Jianchao Ji, Juntao Tan, Shuyuan Xu, Zelong Li, and Yongfeng Zhang. 2023{\natexlab{a}}.
\newblock Open{AGI}: When {LLM} meets domain experts.
\newblock In \emph{Thirty-seventh Conference on Neural Information Processing Systems}.

\bibitem[{Ge et~al.(2023{\natexlab{b}})Ge, Ren, Hua, Xu, Tan, and Zhang}]{ge2023llm}
Yingqiang Ge, Yujie Ren, Wenyue Hua, Shuyuan Xu, Juntao Tan, and Yongfeng Zhang. 2023{\natexlab{b}}.
\newblock Llm as os, agents as apps: Envisioning aios, agents and the aios-agent ecosystem.
\newblock \emph{arXiv}.

\bibitem[{Geng et~al.(2024)Geng, D{\"o}ner, Wendler, Josifoski, and West}]{geng2024sketch}
Saibo Geng, Berkay D{\"o}ner, Chris Wendler, Martin Josifoski, and Robert West. 2024.
\newblock Sketch-guided constrained decoding for boosting blackbox large language models without logit access.
\newblock \emph{arXiv preprint arXiv:2401.09967}.

\bibitem[{Geng et~al.(2023)Geng, Josifoski, Peyrard, and West}]{geng2023grammar}
Saibo Geng, Martin Josifoski, Maxime Peyrard, and Robert West. 2023.
\newblock Grammar-constrained decoding for structured nlp tasks without finetuning.
\newblock In \emph{Proceedings of the 2023 Conference on Empirical Methods in Natural Language Processing}, pages 10932--10952.

\bibitem[{Gu et~al.(2022)Gu, Feng, Ma, Zhang, Gong, and Qin}]{gu-etal-2022-distributional}
Yuxuan Gu, Xiaocheng Feng, Sicheng Ma, Lingyuan Zhang, Heng Gong, and Bing Qin. 2022.
\newblock \href {https://doi.org/10.18653/v1/2022.emnlp-main.67} {A distributional lens for multi-aspect controllable text generation}.
\newblock In \emph{Proceedings of the 2022 Conference on Empirical Methods in Natural Language Processing}, pages 1023--1043, Abu Dhabi, United Arab Emirates. Association for Computational Linguistics.

\bibitem[{Hemmer et~al.(2023)Hemmer, Coustaty, Bartolo, Brachat, and Ogier}]{hemmer2023lazy}
Arthur Hemmer, Mickael Coustaty, Nicola Bartolo, Jerome Brachat, and Jean-Marc Ogier. 2023.
\newblock Lazy-k decoding: Constrained decoding for information extraction.
\newblock In \emph{Proceedings of the 2023 Conference on Empirical Methods in Natural Language Processing}, pages 6727--6736.

\bibitem[{Hessel et~al.(2021)Hessel, Holtzman, Forbes, Bras, and Choi}]{hessel2021clipscore}
Jack Hessel, Ari Holtzman, Maxwell Forbes, Ronan~Le Bras, and Yejin Choi. 2021.
\newblock {CLIPScore:} a reference-free evaluation metric for image captioning.

\bibitem[{Hopcroft et~al.(2001)Hopcroft, Motwani, and Ullman}]{hopcroft2001introduction}
John~E Hopcroft, Rajeev Motwani, and Jeffrey~D Ullman. 2001.
\newblock Introduction to automata theory, languages, and computation.
\newblock \emph{Acm Sigact News}, 32(1):60--65.

\bibitem[{Hu et~al.(2021)Hu, Shen, Wallis, Allen-Zhu, Li, Wang, Wang, and Chen}]{hu2021lora}
Edward~J. Hu, Yelong Shen, Phillip Wallis, Zeyuan Allen-Zhu, Yuanzhi Li, Shean Wang, Lu~Wang, and Weizhu Chen. 2021.
\newblock \href {https://arxiv.org/abs/2106.09685} {Lora: Low-rank adaptation of large language models}.
\newblock \emph{Preprint}, arXiv:2106.09685.

\bibitem[{Hua et~al.(2023)Hua, Fan, Li, Mei, Ji, Ge, Hemphill, and Zhang}]{hua2023war}
Wenyue Hua, Lizhou Fan, Lingyao Li, Kai Mei, Jianchao Ji, Yingqiang Ge, Libby Hemphill, and Yongfeng Zhang. 2023.
\newblock War and peace (waragent): Large language model-based multi-agent simulation of world wars.
\newblock \emph{arXiv:2311.17227}.

\bibitem[{Huang et~al.(2022)Huang, Abbeel, Pathak, and Mordatch}]{huang2022language}
Wenlong Huang, Pieter Abbeel, Deepak Pathak, and Igor Mordatch. 2022.
\newblock Language models as zero-shot planners: Extracting actionable knowledge for embodied agents.
\newblock In \emph{Proceedings of the 39th International Conference on Machine Learning}, volume 162 of \emph{Proceedings of Machine Learning Research}, pages 9118--9147. PMLR.

\bibitem[{Lewis et~al.(2019)Lewis, Liu, Goyal, Ghazvininejad, Mohamed, Levy, Stoyanov, and Zettlemoyer}]{lewis2019bart}
Mike Lewis, Yinhan Liu, Naman Goyal, Marjan Ghazvininejad, Abdelrahman Mohamed, Omer Levy, Ves Stoyanov, and Luke Zettlemoyer. 2019.
\newblock \href {https://arxiv.org/abs/1910.13461} {Bart: Denoising sequence-to-sequence pre-training for natural language generation, translation, and comprehension}.
\newblock \emph{Preprint}, arXiv:1910.13461.

\bibitem[{Li et~al.(2023{\natexlab{a}})Li, Hammoud, Itani, Khizbullin, and Ghanem}]{li2023camel}
Guohao Li, Hasan Abed Al~Kader Hammoud, Hani Itani, Dmitrii Khizbullin, and Bernard Ghanem. 2023{\natexlab{a}}.
\newblock Camel: Communicative agents for "mind" exploration of large language model society.
\newblock In \emph{Thirty-seventh Conference on Neural Information Processing Systems}.

\bibitem[{Li et~al.(2020)Li, Chen, and Zhang}]{li2020towards}
Lei Li, Li~Chen, and Yongfeng Zhang. 2020.
\newblock Towards controllable explanation generation for recommender systems via neural template.
\newblock In \emph{Companion proceedings of the web conference 2020}, pages 198--202.

\bibitem[{Li et~al.(2023{\natexlab{b}})Li, Lv, Chen, Cui, Lu, Florencio, Zhang, Li, and Wei}]{li2023trocr}
Minghao Li, Tengchao Lv, Jingye Chen, Lei Cui, Yijuan Lu, Dinei Florencio, Cha Zhang, Zhoujun Li, and Furu Wei. 2023{\natexlab{b}}.
\newblock Trocr: transformer-based optical character recognition with pre-trained models.
\newblock In \emph{Proceedings of the Thirty-Seventh AAAI Conference on Artificial Intelligence and Thirty-Fifth Conference on Innovative Applications of Artificial Intelligence and Thirteenth Symposium on Educational Advances in Artificial Intelligence}, pages 13094--13102.

\bibitem[{Liu et~al.(2023)Liu, Yu, Zhang, Xu, Lei, Lai, Gu, Ding, Men, Yang, Zhang, Deng, Zeng, Du, Zhang, Shen, Zhang, Su, Sun, Huang, Dong, and Tang}]{liu2023agentbench}
Xiao Liu, Hao Yu, Hanchen Zhang, Yifan Xu, Xuanyu Lei, Hanyu Lai, Yu~Gu, Hangliang Ding, Kaiwen Men, Kejuan Yang, Shudan Zhang, Xiang Deng, Aohan Zeng, Zhengxiao Du, Chenhui Zhang, Sheng Shen, Tianjun Zhang, Yu~Su, Huan Sun, Minlie Huang, Yuxiao Dong, and Jie Tang. 2023.
\newblock \href {https://arxiv.org/abs/2308.03688} {Agentbench: Evaluating llms as agents}.
\newblock \emph{Preprint}, arXiv:2308.03688.

\bibitem[{Lu et~al.(2022)Lu, Welleck, Hessel, Jiang, Qin, West, Ammanabrolu, and Choi}]{lu2022quark}
Ximing Lu, Sean Welleck, Jack Hessel, Liwei Jiang, Lianhui Qin, Peter West, Prithviraj Ammanabrolu, and Yejin Choi. 2022.
\newblock Quark: Controllable text generation with reinforced unlearning.
\newblock \emph{Advances in neural information processing systems}, 35:27591--27609.

\bibitem[{Mei et~al.(2024)Mei, Li, Xu, Ye, Ge, and Zhang}]{mei2024aios}
Kai Mei, Zelong Li, Shuyuan Xu, Ruosong Ye, Yingqiang Ge, and Yongfeng Zhang. 2024.
\newblock Aios: Llm agent operating system.
\newblock \emph{arXiv}.

\bibitem[{OpenAI(2023)}]{openai2023gpt4}
Josh et~al OpenAI. 2023.
\newblock \href {https://arxiv.org/abs/2303.08774} {Gpt-4 technical report}.
\newblock \emph{Preprint}, arXiv:2303.08774.

\bibitem[{Qian et~al.(2023)Qian, Cong, Liu, Yang, Chen, Su, Dang, Li, Xu, Li, Liu, and Sun}]{qian2023communicative}
Chen Qian, Xin Cong, Wei Liu, Cheng Yang, Weize Chen, Yusheng Su, Yufan Dang, Jiahao Li, Juyuan Xu, Dahai Li, Zhiyuan Liu, and Maosong Sun. 2023.
\newblock \href {https://arxiv.org/abs/2307.07924} {Communicative agents for software development}.
\newblock \emph{Preprint}, arXiv:2307.07924.

\bibitem[{Qin et~al.(2022)Qin, Welleck, Khashabi, and Choi}]{qin2022cold}
Lianhui Qin, Sean Welleck, Daniel Khashabi, and Yejin Choi. 2022.
\newblock Cold decoding: Energy-based constrained text generation with langevin dynamics.
\newblock \emph{Advances in Neural Information Processing Systems}, 35:9538--9551.

\bibitem[{Qin et~al.(2023)Qin, Hu, Lin, Chen, Ding, Cui, Zeng, Huang, Xiao, Han, Fung, Su, Wang, Qian, Tian, Zhu, Liang, Shen, Xu, Zhang, Ye, Li, Tang, Yi, Zhu, Dai, Yan, Cong, Lu, Zhao, Huang, Yan, Han, Sun, Li, Phang, Yang, Wu, Ji, Liu, and Sun}]{qin2023tool}
Yujia Qin, Shengding Hu, Yankai Lin, Weize Chen, Ning Ding, Ganqu Cui, Zheni Zeng, Yufei Huang, Chaojun Xiao, Chi Han, Yi~Ren Fung, Yusheng Su, Huadong Wang, Cheng Qian, Runchu Tian, Kunlun Zhu, Shihao Liang, Xingyu Shen, Bokai Xu, Zhen Zhang, Yining Ye, Bowen Li, Ziwei Tang, Jing Yi, Yuzhang Zhu, Zhenning Dai, Lan Yan, Xin Cong, Yaxi Lu, Weilin Zhao, Yuxiang Huang, Junxi Yan, Xu~Han, Xian Sun, Dahai Li, Jason Phang, Cheng Yang, Tongshuang Wu, Heng Ji, Zhiyuan Liu, and Maosong Sun. 2023.
\newblock \href {https://arxiv.org/abs/2304.08354} {Tool learning with foundation models}.
\newblock \emph{Preprint}, arXiv:2304.08354.

\bibitem[{Radford et~al.(2019)Radford, Wu, Child, Luan, Amodei, Sutskever et~al.}]{radford2019language}
Alec Radford, Jeffrey Wu, Rewon Child, David Luan, Dario Amodei, Ilya Sutskever, et~al. 2019.
\newblock Language models are unsupervised multitask learners.
\newblock \emph{OpenAI blog}, 1(8):9.

\bibitem[{Raffel et~al.(2020)Raffel, Shazeer, Roberts, Lee, Narang, Matena, Zhou, Li, and Liu}]{raffel2020exploring}
Colin Raffel, Noam Shazeer, Adam Roberts, Katherine Lee, Sharan Narang, Michael Matena, Yanqi Zhou, Wei Li, and Peter~J Liu. 2020.
\newblock Exploring the limits of transfer learning with a unified text-to-text transformer.
\newblock \emph{The Journal of Machine Learning Research}, 21(1):5485--5551.

\bibitem[{Rombach et~al.(2022)Rombach, Blattmann, Lorenz, Esser, and Ommer}]{rombach2022high}
Robin Rombach, Andreas Blattmann, Dominik Lorenz, Patrick Esser, and Bj{\"o}rn Ommer. 2022.
\newblock High-resolution image synthesis with latent diffusion models.
\newblock In \emph{Proceedings of the IEEE/CVF conference on computer vision and pattern recognition}, pages 10684--10695.

\bibitem[{Sanh et~al.(2020)Sanh, Debut, Chaumond, and Wolf}]{sanh2020distilbert}
Victor Sanh, Lysandre Debut, Julien Chaumond, and Thomas Wolf. 2020.
\newblock \href {https://arxiv.org/abs/1910.01108} {Distilbert, a distilled version of bert: smaller, faster, cheaper and lighter}.
\newblock \emph{Preprint}, arXiv:1910.01108.

\bibitem[{Schick et~al.(2023)Schick, Dwivedi-Yu, Dessì, Raileanu, Lomeli, Zettlemoyer, Cancedda, and Scialom}]{schick2023toolformer}
Timo Schick, Jane Dwivedi-Yu, Roberto Dessì, Roberta Raileanu, Maria Lomeli, Luke Zettlemoyer, Nicola Cancedda, and Thomas Scialom. 2023.
\newblock \href {https://arxiv.org/abs/2302.04761} {Toolformer: Language models can teach themselves to use tools}.
\newblock \emph{Preprint}, arXiv:2302.04761.

\bibitem[{Takase and Okazaki(2019)}]{takase2019positional}
Sho Takase and Naoaki Okazaki. 2019.
\newblock Positional encoding to control output sequence length.
\newblock In \emph{Proceedings of the 2019 Conference of the North}. Association for Computational Linguistics.

\bibitem[{Touvron et~al.(2023{\natexlab{a}})Touvron, Lavril, Izacard, Martinet, Lachaux, Lacroix, Rozi{\`e}re, Goyal, Hambro, Azhar et~al.}]{touvron2023llama}
Hugo Touvron, Thibaut Lavril, Gautier Izacard, Xavier Martinet, Marie-Anne Lachaux, Timoth{\'e}e Lacroix, Baptiste Rozi{\`e}re, Naman Goyal, Eric Hambro, Faisal Azhar, et~al. 2023{\natexlab{a}}.
\newblock \href {https://arxiv.org/abs/2302.13971} {Llama: Open and efficient foundation language models}.
\newblock \emph{Preprint}, arXiv:2302.13971.

\bibitem[{Touvron et~al.(2023{\natexlab{b}})Touvron, Martin, Stone, Albert, Almahairi, Babaei, Bashlykov, Batra, Bhargava, Bhosale et~al.}]{touvron2023llama2}
Hugo Touvron, Louis Martin, Kevin Stone, Peter Albert, Amjad Almahairi, Yasmine Babaei, Nikolay Bashlykov, Soumya Batra, Prajjwal Bhargava, Shruti Bhosale, et~al. 2023{\natexlab{b}}.
\newblock \href {https://arxiv.org/abs/2307.09288} {Llama 2: Open foundation and fine-tuned chat models}.
\newblock \emph{Preprint}, arXiv:2307.09288.

\bibitem[{Wang et~al.(2022)Wang, Yang, Hu, Li, Lin, Gan, Liu, Liu, and Wang}]{wang2022git}
Jianfeng Wang, Zhengyuan Yang, Xiaowei Hu, Linjie Li, Kevin Lin, Zhe Gan, Zicheng Liu, Ce~Liu, and Lijuan Wang. 2022.
\newblock \href {https://arxiv.org/abs/2205.14100} {Git: A generative image-to-text transformer for vision and language}.
\newblock \emph{Preprint}, arXiv:2205.14100.

\bibitem[{Wang et~al.(2023)Wang, Ma, Feng, Zhang, Yang, Zhang, Chen, Tang, Chen, Lin, Zhao, Wei, and Wen}]{wang2023survey}
Lei Wang, Chen Ma, Xueyang Feng, Zeyu Zhang, Hao Yang, Jingsen Zhang, Zhiyuan Chen, Jiakai Tang, Xu~Chen, Yankai Lin, Wayne~Xin Zhao, Zhewei Wei, and Ji-Rong Wen. 2023.
\newblock \href {https://arxiv.org/abs/2308.11432} {A survey on large language model based autonomous agents}.
\newblock \emph{Preprint}, arXiv:2308.11432.

\bibitem[{Williams(1992)}]{williams1992simple}
Ronald~J Williams. 1992.
\newblock Simple statistical gradient-following algorithms for connectionist reinforcement learning.
\newblock \emph{Machine learning}, 8:229--256.

\bibitem[{Wu et~al.(2020)Wu, Xu, Dai, Wan, Zhang, Yan, Tomizuka, Gonzalez, Keutzer, and Vajda}]{wu2020visual}
Bichen Wu, Chenfeng Xu, Xiaoliang Dai, Alvin Wan, Peizhao Zhang, Zhicheng Yan, Masayoshi Tomizuka, Joseph Gonzalez, Kurt Keutzer, and Peter Vajda. 2020.
\newblock \href {https://arxiv.org/abs/2006.03677} {Visual transformers: Token-based image representation and processing for computer vision}.
\newblock \emph{Preprint}, arXiv:2006.03677.

\bibitem[{Xi et~al.(2023)Xi, Chen, Guo, He, Ding, Hong, Zhang, Wang, Jin, Zhou, Zheng, Fan, Wang, Xiong, Zhou, Wang, Jiang, Zou, Liu, Yin, Dou, Weng, Cheng, Zhang, Qin, Zheng, Qiu, Huang, and Gui}]{xi2023rise}
Zhiheng Xi, Wenxiang Chen, Xin Guo, Wei He, Yiwen Ding, Boyang Hong, Ming Zhang, Junzhe Wang, Senjie Jin, Enyu Zhou, Rui Zheng, Xiaoran Fan, Xiao Wang, Limao Xiong, Yuhao Zhou, Weiran Wang, Changhao Jiang, Yicheng Zou, Xiangyang Liu, Zhangyue Yin, Shihan Dou, Rongxiang Weng, Wensen Cheng, Qi~Zhang, Wenjuan Qin, Yongyan Zheng, Xipeng Qiu, Xuanjing Huang, and Tao Gui. 2023.
\newblock \href {https://arxiv.org/abs/2309.07864} {The rise and potential of large language model based agents: A survey}.
\newblock \emph{Preprint}, arXiv:2309.07864.

\bibitem[{Xie et~al.(2024)Xie, Zhang, Chen, Zhu, Lou, Tian, Xiao, and Su}]{xie2024travelplanner}
Jian Xie, Kai Zhang, Jiangjie Chen, Tinghui Zhu, Renze Lou, Yuandong Tian, Yanghua Xiao, and Yu~Su. 2024.
\newblock Travelplanner: A benchmark for real-world planning with language agents.
\newblock \emph{arXiv preprint arXiv:2402.01622}.

\bibitem[{Yao et~al.(2023)Yao, Zhao, Yu, Du, Shafran, Narasimhan, and Cao}]{yao2023react}
Shunyu Yao, Jeffrey Zhao, Dian Yu, Nan Du, Izhak Shafran, Karthik Narasimhan, and Yuan Cao. 2023.
\newblock React: Synergizing reasoning and acting in language models.
\newblock In \emph{International Conference on Learning Representations (ICLR)}.

\bibitem[{Yuan et~al.(2024)Yuan, Song, Chen, Tan, Shen, Kan, Li, and Yang}]{yuan2024easytool}
Siyu Yuan, Kaitao Song, Jiangjie Chen, Xu~Tan, Yongliang Shen, Ren Kan, Dongsheng Li, and Deqing Yang. 2024.
\newblock Easytool: Enhancing llm-based agents with concise tool instruction.
\newblock \emph{arXiv preprint arXiv:2401.06201}.

\bibitem[{Zamir et~al.(2022)Zamir, Arora, Khan, Hayat, Khan, and Yang}]{zamir2022restormer}
Syed~Waqas Zamir, Aditya Arora, Salman Khan, Munawar Hayat, Fahad~Shahbaz Khan, and Ming-Hsuan Yang. 2022.
\newblock Restormer: Efficient transformer for high-resolution image restoration.
\newblock In \emph{Proceedings of the IEEE/CVF conference on computer vision and pattern recognition}, pages 5728--5739.

\bibitem[{Zhang et~al.(2017)Zhang, Zhu, Isola, Geng, Lin, Yu, and Efros}]{zhang2017real}
Richard Zhang, Jun-Yan Zhu, Phillip Isola, Xinyang Geng, Angela~S Lin, Tianhe Yu, and Alexei~A Efros. 2017.
\newblock Real-time user-guided image colorization with learned deep priors.
\newblock \emph{ACM Transactions on Graphics (TOG)}, 36(4):1--11.

\bibitem[{Zhang et~al.(2020)Zhang, Kishore, Wu, Weinberger, and Artzi}]{bert-score}
Tianyi Zhang, Varsha Kishore, Felix Wu, Kilian~Q. Weinberger, and Yoav Artzi. 2020.
\newblock Bertscore: Evaluating text generation with bert.

\end{thebibliography}

\onecolumn
\appendix
\newpage

\footnotetext[1]{\label{colorizer}\url{https://github.com/richzhang/colorization}}
\footnotetext[2]{\label{resolution}\url{https://huggingface.co/caidas/swin2SR-classical-sr-x2-64}}
\footnotetext[3]{\label{restormer}\url{https://github.com/swz30/Restormer}}
\footnotetext[4]{\label{ViT}\url{https://huggingface.co/google/vit-base-patch16-224}}
\footnotetext[5]{\label{DETR}\url{https://huggingface.co/facebook/detr-resnet-101}}
\footnotetext[6]{\label{caption}\url{https://huggingface.co/nlpconnect/vit-gpt2-image-captioning}}
\footnotetext[7]{\label{diffusion}\url{https://huggingface.co/CompVis/stable-diffusion-v1-4}}
\footnotetext[8]{\label{sentimenta}\url{https://huggingface.co/distilbert-base-uncased-finetuned-sst-2-english}}
\footnotetext[9]{\label{BART}\url{https://huggingface.co/facebook/bart-large-cnn}}
\footnotetext[10]{\label{T5}\url{https://huggingface.co/t5-base}}
\footnotetext[11]{\label{mask}\url{https://huggingface.co/distilbert-base-uncased}}
\footnotetext[12]{\label{gpt2}\url{https://huggingface.co/gpt2}}
\footnotetext[13]{\label{GIT}\url{https://huggingface.co/microsoft/git-base-textvqa}}
\footnotetext[14]{\label{QA}\url{https://huggingface.co/distilbert-base-cased-distilled-squad}}

\section{Appendix}
\label{sec:appendix}

\subsection{OpenAGI Benchmark Tasks and Tools}
\label{sec:task_example}

The tool list integrated in the OpenAGI platform \citep{openagi} is shown in Table \ref{Table:openagi_tool} and the different types of tasks as well as example tasks are shown in Table \ref{Table:task_example}.

\begin{table}[ht]
\small
    \centering
    \setlength{\tabcolsep}{0.5pt}
    \begin{tabular}{c|c|l|l}
    \hline
      Input Modality & Output Modality & Tool Description & Expert Model \\
      \hline
      \multirow{7.5}{*}{Image} & \multirow{4}{*}{Image ($A$)} & Colorization ($a_1$) & Colorizer\footref{colorizer} \citep{zhang2017real} \\
      & & Super-Resolution ($a_2$) & Swin2SR\footref{resolution} \citep{conde2022swin2sr} \\
      & & Image Denoising ($a_3$) & Restormer\footref{restormer} \citep{zamir2022restormer}\\
      & & Image Deblurring ($a_4$) & Restormer\footref{restormer} \citep{zamir2022restormer}\\
      \cline{2-4}
      & \multirow{3}{*}{Text ($B$)} & Image Classification ($b_1$) & ViT\footref{ViT} \citep{dosovitskiy2020vit}\\
      & & Object Detection ($b_2$) & DETR\footref{DETR} \citep{carion2020end}\\
      & & Image Captioning ($b_3$) & Vision Encoder Decoder\footref{caption} \citep{ li2023trocr}\\
      \hline
      \multirow{6.5}{*}{Text} & Image ($C$) & Text-to-Image Generation ($c_1$) & Stable Diffusion\footref{diffusion} \citep{rombach2022high}\\
      \cline{2-4}
      & \multirow{5}{*}{Text ($D$)} & Sentiment Analysis  ($d_1$)& DistilBERT\footref{sentimenta} \citep{sanh2020distilbert}\\
      & & Text Summarization ($d_2$) & BART\footref{BART} \citep{lewis2019bart}\\
      & & Machine Translation ($d_3$) & T5\footref{T5} \citep{raffel2020exploring}\\
      & & Fill Mask ($d_4$) & DistilBERT\footref{mask} \citep{sanh2020distilbert}\\
      & & Text Generation ($d_5$) & GPT-2\footref{gpt2} \citep{radford2019language}\\
      \hline
      Image-Text Pair & Text ($E$) & Visual Question Answering ($e_1$) & GIT\footref{GIT} \citep{wang2022git}\\
      \hline
      Text-Text Pair & Text ($F$) & Question Answering ($f_1$)& DistilBERT\footref{QA} \citep{sanh2020distilbert}\\
      \hline
    \end{tabular}
    \caption{Tool list integrated in OpenAGI platform \citep{openagi} for benchmark tasks. The tools are categorized into six primary groups according to the input and output modalities.}
    \label{Table:openagi_tool}
\end{table}

\subsection{Implementation Details}
\label{sec:implementation}

Our framework and all baselines are implemented by PyTorch, an open-source library. We follow the implementation setting of the OpenAGI platform \citep{openagi} for baselines. For the F-LLM+RLTF framework, we use the REINFORCE \citep{williams1992simple} as the core reinforcement learning algorithm of RLTF. We use the original checkpoint of each backbone LLM without supervised fine-tuning. We set the maximum number of updating epochs at 30 and use Adam as the optimizer with the learning rate at 0.001 for RLTF. Also, we apply Low-Rank Adaptation (LoRA) \citep{hu2021lora} to the RLTF for efficient fine-tuning with the rank as 8 and the placement as $v_{proj}$ ($W_v$) and $q_{proj}$ ($W_q$). Our experimental results are the average of five runs.

\subsection{Prompt for OpenAGI Benchmark Tasks}
\label{sec:agi_prompt}

An example of the CFG and the corresponding PDA are shown in Eq.\eqref{Eq:agi_cfg} and Eq.\eqref{Eq:agi_cfg_second} as well as Figure \ref{fig:agi_cfg}. Different benchmark tasks share a common subset of constraints shown in Eq.\eqref{Eq:agi_cfg} corresponding to the input-output constraints of the tools. Besides, each task has its own constraint on the input-output of the task, resulting in different constraints in Eq.\eqref{Eq:agi_cfg_second}, and thus the final PDA (Figure \ref{fig:agi_cfg}) for different tasks are different. Also, we directly use the tool name (e.g., \textit{Colorization}) instead of the category name (e.g., ``Image-in, Image-out tools'') in the prompt. For example, consider the PDA in Figure \ref{fig:agi_cfg} when it is in state $q_1$ with the symbol $I$ at the top of the stack. There are three viable transits: $(\varepsilon, I; AI)$, $(\varepsilon, I; CT)$, and $(i, I; \varepsilon)$. In practice, as \textit{nonterminals} $A$ and $C$ are promptly replaced by $terminals$ from $a_1\sim a_4$ and $c_1$. This is because the tool name provides more concrete information for the LLM to comprehend the tool's functionality, enabling the LLM to make informed decisions without influencing the plan's executability.

Zero-shot Prompt:
\begin{lstlisting}[language=HTML]
 Problem: {task_description}.
 What is its soltuion? Use 'Setp' to mark.
\end{lstlisting}

Few-shot Prompt:
\begin{lstlisting}[language=HTML]
 {few shot examples in the format of:
    {
        Problem: {task_description}.
        Solution:
        Step 1: ...
        Step 2: ...
        ...
        Step k: ...
    }
 }
 Problem: {task_description}.
 Solution: 
\end{lstlisting}

RLTF Prompt (RLTF executes the solution and use the performance as reward to fine-tune the LLM):
\begin{lstlisting}[language=HTML]
 Problem: {task_description}.
 Solution: 
\end{lstlisting}

Formal-LLM Prompt (zero-shot):
\begin{lstlisting}[language=HTML]
You will help me generate a plan for the problem: {task_description} by answering a series of my questions.

{current_progress}

To get the {data_modality}, we have the following choices:

{choice_list}

Your answer should be only an integer, referring to the desired choice.
\end{lstlisting}

Formal-LLM + RLTF (execute the solution of Formal-LLM and use the performance as reward to fine-tune the LLM):
\begin{lstlisting}[language=HTML]
You will help me generate a plan for the problem: {task_description} by answering a series of my questions.

{current_progress}

To get the {data_modality}, we have the following choices:

{choice_list}

Your answer should be only an integer, referring to the desired choice.
\end{lstlisting}

Formal-LLM Prompt Example:
\begin{lstlisting}[language=HTML]
You will help me generate a plan for the problem: "Given a grayscale image, how to return the regular image step by step?" by answering a series of my questions.

Current Progress: 

Step n: Use Image Super Resolution;
Step (n-1): ?

To get the input image of "Image Super Resolution", we have the following choices:

1: the output of Colorization,
2: the output of Image Denoising,
3: the output of Image Deblurring,
4: the output of Text to Image Generation,
5: Input Image.

Your answer should be only an integer, referring to the desired choice.
\end{lstlisting}

\begin{table}[t]
\small
    \centering
    \setlength{\tabcolsep}{2pt}
    \begin{tabular}{l|l|l|l|l|l}
    \hline
      Task & Metrics & Output & Label & Evaluation & Task Example \\
      \hline
      \multirow{3}{*}{Task 1} & \multirow{3}{*}{CLIP Score} & \multirow{3}{*}{Image} & \multirow{3}{*}{Text} & \multirow{3}{*}{Text-to-Image similarity} & Given clozed English text, \\&&&&&how to generate an image \\&&&&&step by step? \\
      \hline
      \multirow{3}{*}{Task 2} & \multirow{3}{*}{BERT Score} & \multirow{3}{*}{Text} & \multirow{3}{*}{Text} & \multirow{3}{*}{Text-to-Text similarity} & Given noisy grayscale image, \\&&&&&how to return the caption in \\&&&&&German step by step? \\
      \hline
      \multirow{3}{*}{Task 3} & \multirow{3}{*}{ViT Score} & \multirow{3}{*}{Image} & \multirow{3}{*}{Image} & \multirow{3}{*}{Image-to-Image similarity} & Given blurry grayscale image, \\&&&&&how to return the regular \\&&&&&image step by step? \\
      \hline\hline
      \multirow{5}{*}{Task X} & \multirow{5}{*}{Corresponding Score} & \multirow{5}{*}{/} & \multirow{5}{*}{/} & \multirow{5}{*}{/} & Given low-resolution noisy \\&&&&&blurry grayscale image and \\&&&&&English query, how to \\&&&&&answer the question in \\&&&&&German step by step?\\
      \hline
    \end{tabular}
    \caption{Benchmark task examples under each category. Task X is a subset of ``Task 1 $\cup$ Task 2 $\cup$ Task 3'', which is a subset of tasks that require a tree-structured plan rather than chain-structured plan due to the use of many input-single-output tools. Task X is used to test the complex planning ability of our Formal-LLM framework.
    }
    \label{Table:task_example}
\end{table}

\subsection{Formal-LLM on TravelPlanner Benchmark}
\label{sec:travel_planner}

We apply our Formal-LLM framework to the TravelPlanner \citep{xie2024travelplanner} platform. TravelPlanner is a benchmark designed to assess language agents' performance in tool use and complex planning under various constraints. We use GPT-4 as the backbone LLM because TravelPlanner's experiments show that other LLMs cannot generate valid plans that meet all constraints. We use the workflow in Figure \ref{fig:travel_planner} as the automaton to control the agent planning in the Formal-LLM framework. The flowchart is essentially a Deterministic Finite Automaton (DFA), which is a special case of Pushdown
Automaton (PDA). 
We provide an introduction to the experiment in the following.

\subsubsection{Evaluation Metrics}

\begin{itemize}
    \item \textbf{Delivery Rate}: This metric assesses whether agents can successfully deliver a final plan.
    \item \textbf{Commonsense Constraint Pass Rate}: This metric evaluates whether a plan satisfies the pre-defined eight commonsense constraints.
    \item \textbf{Hard Constraint Pass Rate}: This metric measures whether a plan meets all explicit constraints in the given query.
    \item \textbf{Final Pass Rate}: This metric represents the proportion of the plans that meet all aforementioned constraints among all tested plans.
\end{itemize}

Besides, for the \textbf{Commonsense Constraint Pass Rate} and \textbf{Hard Constraint Pass Rate}, we report the \textbf{Micro Pass Rate}, calculating the ratio of satisfied constraints to the total number of constraints, and the \textbf{Macro Pass Rate}, calculating the ratio of plans that meet all commonsense and hard constraints among all tested plans.

\subsubsection{Baseline}

We utilize GPT-4 as the backbone LLM and adopt the ReAct \citep{yao2023react} framework as our baseline setting, as it demonstrated the highest performance in previous experiments \citep{xie2024travelplanner}.

\begin{figure}[t]
  \centering
  \includegraphics[width=\textwidth]{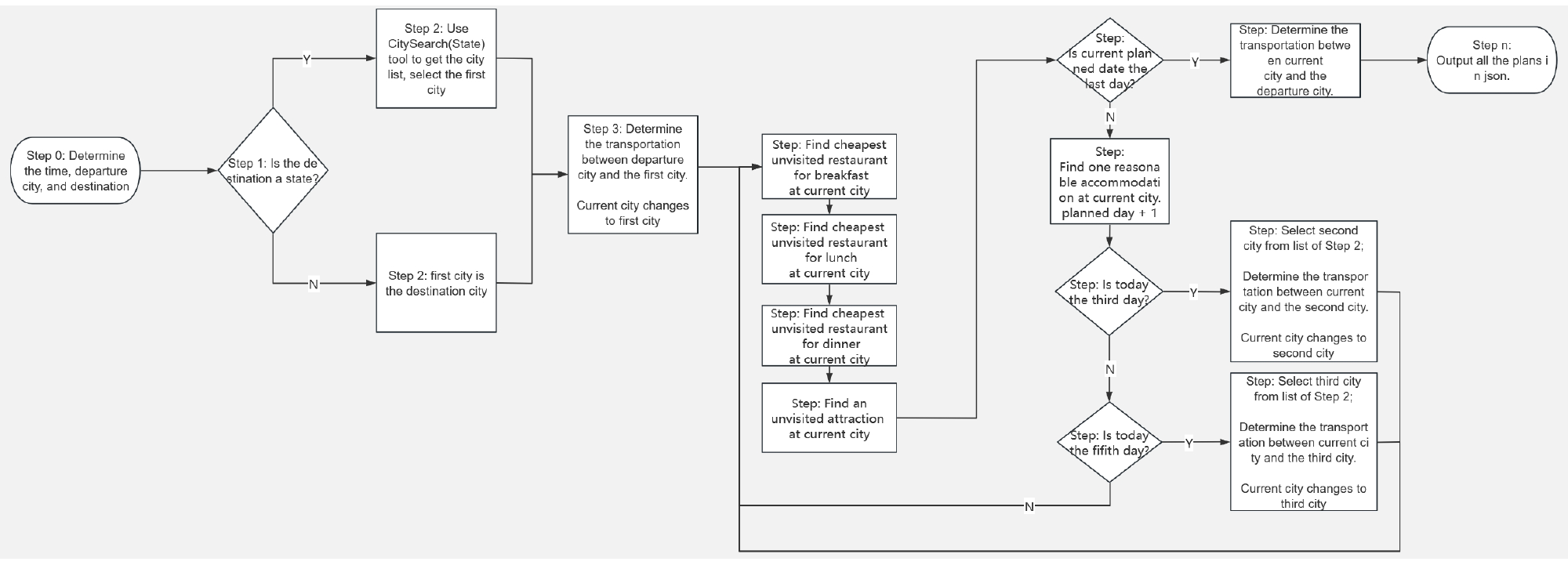}
  \caption{The flowchart for TravelPlanner benchmark.}
  \label{fig:travel_planner}
\end{figure}

\begin{table}[t]
\small
    \centering
    \setlength{\tabcolsep}{4pt}
    \begin{tabular}{c|c|cc|cc|c}
    \hline
    \multirow{1.5}{*}{Planning Strategy} & \multirow{1.5}{*}{Delivery Rate} & \multicolumn{2}{c|}{\shortstack{Commonsense\\Pass Rate}} & \multicolumn{2}{c|}{\shortstack{Hard Constraint\\Pass Rate}} & \multirow{1.5}{*}{Final Pass Rate} \\
    \cline{3-4}\cline{5-6}
    & & Micro & Macro & Micro & Macro \\
    \hline
    ReAct \citep{yao2023react} & 93.1 & 63.3 & 2.0 & 10.5 & 5.5 & 0.6 \\
    Formal-LLM (Ours) & \bm{$100$} & \bm{$82.9$} & \bm{$16.7$} & \bm{$41.3$} & \bm{$31.7$} & \bm{$9.2$} \\
    \hline
    \end{tabular}
    \caption{Experiment results on the testing dataset of TravelPlanner \citep{xie2024travelplanner} benchmark when using GPT-4 as the backbone LLM. The best results are marked in bold.}
    \label{Table:travel_planner}
\end{table}

\subsubsection{Experiment Result}

The experiment results on the TravelPlanner benchmark are shown in Table \ref{Table:travel_planner}. Each row stands for a planning strategy and each column represents an evaluation metric. The results show that our Formal-LLM is better than the best baseline in every metric, validating that our Formal-LLM can generate more reasonable plans. Also, unlike the experiments on the OpenAGI platform, Formal-LLM does not generate 100\% valid plans on the TravelPlanner benchmark, due to the task complexity and automaton quality. However, the generated plans meet the constraints described in the automaton in Figure \ref{fig:travel_planner}. We can improve the automaton quality for TravelPlanner and report even better performances in the future.

\subsection{Prompt for Real-life Tasks}
\label{sec:real_prompt}

\subsubsection{Daily Plan}
\label{sec:daily}
The automaton is displayed in Figure \ref{fig:daily}.

GPT-4 Prompt:
\begin{lstlisting}[language=HTML]
Generate a plan for activities between 10:00 and 20:00.

Breakfast, lunch, and supper need 1 hour each.

Outdoor activities: basketball playing 13:00 - 15:00; do grocery shopping needs 1 hour.

Indoor activities: house cleaning needs 1 hour; homework needs three hours; turning on the washer/laundry machine needs 0 minutes but needs to stay home for one hour.

Other constrain:
Cannot play basketball within an hour after a meal.
\end{lstlisting}

Formal-LLM Prompt:
\begin{lstlisting}[language=HTML]
Generate a plan for activities between 10:00 and 20:00.

Breakfast, lunch, and supper need 1 hour each.

Outdoor activities: basketball playing 13:00 - 15:00; do grocery shopping needs 1 hour.

Indoor activities: house cleaning needs 1 hour; homework needs three hours; turning on the washer/laundry machine needs 0 minutes but needs to stay home for one hour.

Other constrain:
Cannot play basketball within an hour after a meal.

Let's start planning from the end.

{current_progress}

Decide on the activity ending at {current_hour}:00.

Here are possible options:

{choice_list}

Your reply should be only one number, such as 1, referring to the option.
\end{lstlisting}

Formal-LLM Prompt Example:

\begin{lstlisting}[language=HTML]
Generate a plan for activities between 10:00 and 20:00. 

Breakfast, lunch, and supper need 1 hour each.

Outdoor activities: basketball playing 13:00 - 15:00; do grocery shopping needs 1 hour.

Indoor activities: house cleaning needs 1 hour; homework needs three hours; turning on the washer/laundry machine needs 0 minutes but needs to stay home for one hour.

Other constrain:
Cannot play basketball within an hour after a meal.

Let's start planning from the end.

Current Progress:

17:00 - 20:00 Doing homework.

Decide on the activity ending at 17:00.

Here are possible options:

1. Eating supper.
2. Grocery shopping.
3. House cleaning.
4. Turning on the washer/laundry machine.
5. Doing nothing for one hour.

Your reply should be only one number, such as 1, referring to the option.
\end{lstlisting}

\subsubsection{Cooking Recipe}
\label{sec:cooking}

The formal language for the cooking task is a CFG, shown as Eq.\eqref{eq:cooking}.
\begin{equation}
\begin{split}
& S \rightarrow aAB \\
& B \rightarrow aCD | aCE\\
& C \rightarrow aF \\
& D \rightarrow aAI \\
& E \rightarrow bH \\
& F \rightarrow bAG \\
& G \rightarrow cd \\
& H \rightarrow ef | eI \\
& I \rightarrow cf
\end{split}
\label{eq:cooking}
\end{equation}
where nonterminals: $S$ is for the final beef broccoli; $A$ is for seasoning and can be replaced by an element of the power set of $\{$onion, ginger, garlic, oil, salt, light soy oil, cooking wine, white pepper, sugar, vinegar$\}$; $B$ is for beef and broccoli mixture; $C$ is for lightly cooked beef; $D$ is for lightly cooked broccoli; $E$ is for blanched and drained broccoli; $F$ is for marinaded beef; $G$ is for clean slices of beef; $H$ is for blanched broccoli; $I$ is for clean broccoli; and terminals: $a$ is for wok; $b$ is for bowl; $c$ is for water; $d$ is for raw beef slices; $e$ is for cooking pot; $f$ is for broccoli.

GPT-4 Prompt:
\begin{lstlisting}[language=HTML]
Generate a broccoli beef cooking plan.

The ingredients include raw beef slices, carpaccio, broccoli, onions, ginger, garlic, and water.

The seasonings include cooking oil, salt, light soy sauce, cooking wine, white pepper, sugar, vinegar.

Cooking utensils including woks and cooking pots.

Tableware including chopsticks, spoons, wok spoons, and several bowls.
\end{lstlisting}

Formal-LLM Prompt:

\begin{lstlisting}[language=HTML]
Generate a broccoli beef cooking plan.

The ingredients include raw beef slices, carpaccio, broccoli, onions, ginger, garlic, and water.

The seasonings include cooking oil, salt, light soy sauce, cooking wine, white pepper, sugar, vinegar.

Cooking utensils including woks and cooking pots.

Tableware including chopsticks, spoons, wok spoons, and several bowls.

{current_progress}

Decide on the previous step before current progress.

Here are possible options to get {target_item} for the step: {parent_step}.

{choice_list}

Your reply should be only one number, such as 1, referring to the option.
\end{lstlisting}

Formal-LLM Prompt Example:

\begin{lstlisting}[language=HTML]
Generate a broccoli beef cooking plan.

The ingredients include raw beef slices, carpaccio, broccoli, onions, ginger, garlic, and water.

The seasonings include cooking oil, salt, light soy sauce, cooking wine, white pepper, sugar, vinegar.

Cooking utensils including woks and cooking pots.

Tableware including chopsticks, spoons, wok spoons, and several bowls.

Current Progress:

Step n: Then, we get the cooked broccoli beef.
Step n-1: Stir-fry the beef and broccoli mixture with the seasoning in a wok.
Step n-2: Prepare the seasoning: ginger, garlic, cooking oil, salt, light soy sauce, cooking wine, white pepper for the step: "Stir-fry the beef and broccoli mixture with the seasoning in a wok."
Step n-3: ?

Decide on the previous step before current progress.

Here are possible options to get the mixture of beef and broccoli for the step: "Stir-fry the beef and broccoli mixture with the seasoning in a wok."

1: Combine lightly cooked beef and lightly cooked broccoli in a wok.
2: Combine lightly cooked beef and blanched and drained broccoli in a wok.

Your reply should be only one number, such as 1, referring to the option.
\end{lstlisting}

\subsubsection{Risk Management}
\label{sec:risk}

\begin{figure}[t]
  \centering
  \includegraphics[width=0.9\textwidth]{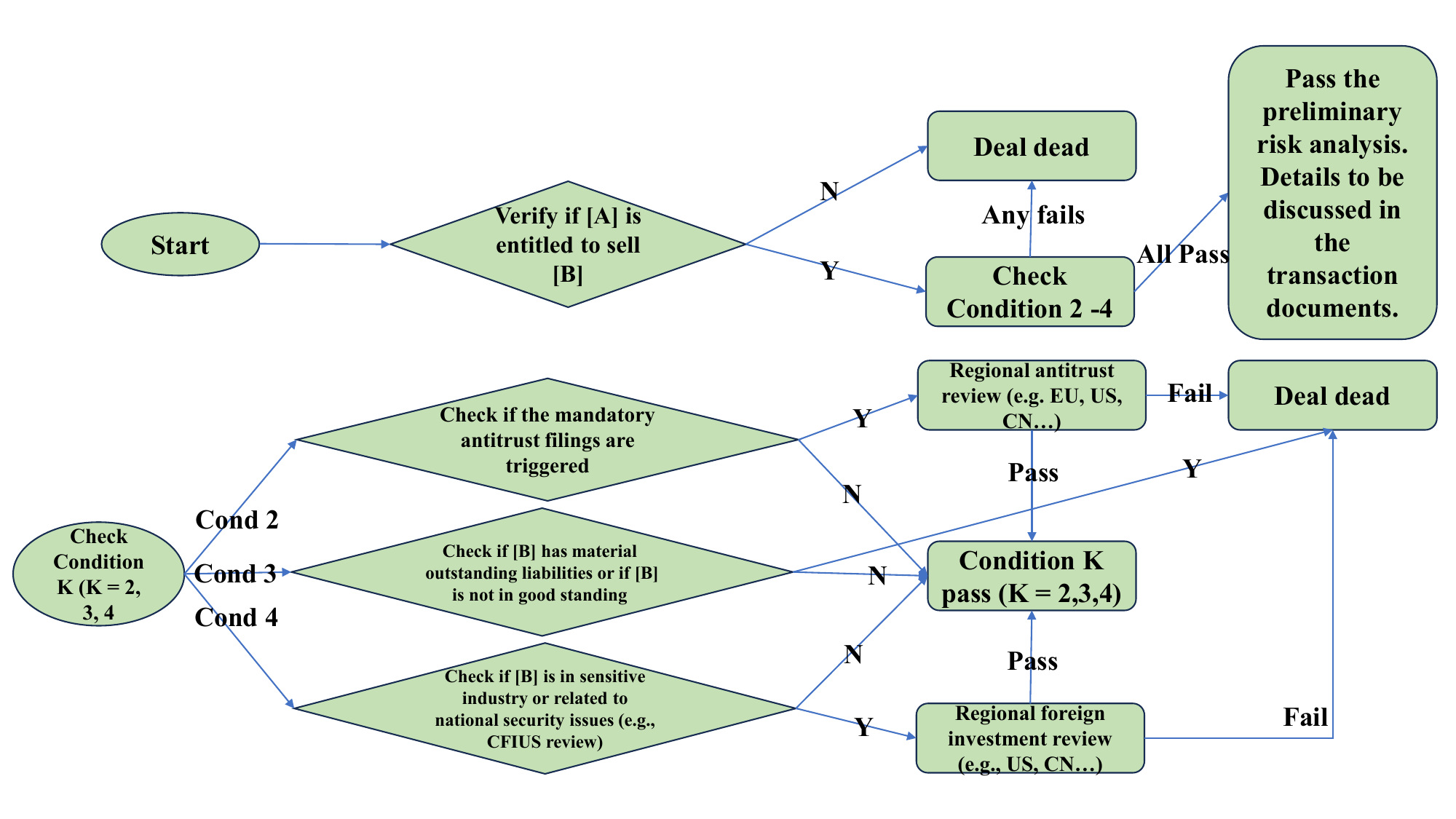}
  \vspace{-10pt}
  \caption{The flowchart for risk management task.}
  \label{fig:risk_management}
  \vspace{-10pt}
\end{figure}

Domain experts draw a flowchart of risk management as Figure \ref{fig:risk_management}. The flowchart is essentially a Deterministic Finite Automaton (DFA), which is a special case of Pushdown Automaton (PDA). A question is generated to inquiry LLM when the automaton reaches a new state. The LLM can select a reasonable branch according to its knowledge of the specific seller, buyer, and company name. In the experiment, the seller is \textit{Shareholders of Blizzard Entertainment}, the company is \textit{Blizzard Entertainment}, and the buyer is \textit{Microsoft}.

GPT-4 Prompt:
\begin{lstlisting}[language=HTML]
Task: You are a plan maker to design a risk management plan for deals related to companies.

Scenario: {Seller A} wants to sell the {Company B} which {Buyer C} wants to buy. Design a risk management plan for {Buyer C} before negotiating a detailed contract.
\end{lstlisting}

Formal-LLM Prompt:
\begin{lstlisting}[language=HTML]
Task: You are a plan maker to design a risk management plan for deals related to companies.

Scenario: {Seller A} wants to sell the {Company B} which {Buyer C} wants to buy. Design a risk management plan for {Buyer C} before negotiating a detailed contract.

{question in the flowchart}

{choice_list}

Your answer should be only one number, such as 1, referring to the option.
\end{lstlisting}

Formal-LLM Prompt Example:
\begin{lstlisting}[language=HTML]
Task: You are a plan maker to design a risk management plan for deals related to companies.

Scenario: Shareholders of Blizzard Entertainment wants to sell the Blizzard Entertainment which Microsoft wants to buy. Design a risk management plan for Microsoft before negotiating a detailed contract.

Question: Does Microsoft buying Blizzard Entertainment trigger mandatory antitrust filing?

1: Yes.
2: No.
3: Insufficient information to make a judgment.

Your answer should be only one number, such as 1, referring to the option.
\end{lstlisting}

\subsection{Complete Results of Real-life Tasks}
\label{sec:real_world_result}

Daily plan result from GPT-4:
\begin{lstlisting}[language=HTML]
10:00 - 11:00 : Breakfast
11:00 - 12:00 : House Cleaning
12:00 - 13:00 : Free Time
13:00 - 15:00 : Basketball Playing
15:00 - 16:00 : Free Time
16:00 - 17:00 : Lunch
17:00 - 18:00 : Grocery Shopping
18:00 - 19:00 : Homework 
19:00 - 20:00 : Laundry and Supper 
20:00 - 23:00 : Continue Homework 

In this plan, basketball playing is scheduled from 13.00 to 15.00 with an hour after breakfast to avoid playing right after eating as per the constraint. Lunch is then scheduled after basketball and Shopping done after that. Since laundry doesn't take time, it has been added with supper and rest of the time is allocated to homework.
\end{lstlisting}

Daily plan result after applying Formal-LLM:
\begin{lstlisting}[language=HTML]
10:00 - 10:00 Turning on the washer/laundry machine.
10:00 - 11:00 House cleaning.
11:00 - 12:00 Eating breakfast.
12:00 - 13:00 Grocery shopping.
13:00 - 15:00 Playing basketball.
15:00 - 16:00 Eating lunch.
16:00 - 19:00 Doing homework.
19:00 - 20:00 Eating supper.
\end{lstlisting}

Cooking recipe result from GPT-4:
\begin{lstlisting}[language=HTML]
Ingredients Preparation
1. Sliced Beef: 300 grams
2. Chinese Broccoli (Gai Lan): As needed, cut into segments
3. Scallions: 1 stalk, cut into sections
4. Ginger: As needed, sliced
5. Garlic: As needed, minced
6. Water: As needed
7. Cooking Oil: As needed
8. Salt: As needed
9. Light Soy Sauce: 2 tablespoons
10. Cooking Wine: 1 tablespoon
11. White Pepper Powder: As needed
12. Sugar: 1 tablespoon
13. Vinegar: 1 tablespoon

Cooking Utensils and Tableware Preparation
1. Stir-fry Pan: 1
2. Cooking Pot: 1
3. Tableware: Chopsticks, Soup Spoon, Stir-fry Pan Spoon, Several Bowls

Cooking Steps

Preparatory Steps
1. Marinate the beef slices with salt and cooking wine, stirring evenly, and let it marinate for 10 minutes.
2. Heat the stir-fry pan and add an appropriate amount of cooking oil.

Sautéing the Beef
1. Once the oil is hot, quickly stir-fry the marinated beef slices until they change color.
2. Add scallion sections, ginger slices, and minced garlic. Continue stirring until a fragrant aroma is released.

Sautéing the Chinese Broccoli
1. In the stir-fry pan, add water. Once the water boils, add the Chinese broccoli.
2. Quickly stir-fry the Chinese broccoli until it changes color, and season with salt.

Seasoning
1. Add light soy sauce, white pepper powder, sugar, and vinegar. Stir well.
2. Continue stirring to ensure that the seasoning is evenly coated on the beef and Chinese broccoli.

Completion
1. Ensure the beef is cooked through, and the Chinese broccoli is flavorful. Remove from heat.

Serving
1. Transfer the Chinese Broccoli Beef to bowls.
2. Serve with rice or noodles.

Tips
1. Avoid marinating the beef slices for too long to maintain a good texture.
2. Control the sautéing time for Chinese broccoli to preserve its crisp and tender texture.
3. Adjust the seasoning according to personal taste preferences, adding or reducing the amount of salt, sugar, and vinegar.
\end{lstlisting}

Cooking recipe result after applying Formal-LLM:
\begin{lstlisting}[language=HTML]
Here is the final plan:

Step 1: Wash raw beef slices with water.
Step 2: Prepare the seasoning: ginger, garlic, cooking oil, salt, light soy sauce, cooking wine, white pepper for the step: "Marinate clean slices of beef in a bowl with seasoning".
Step 3: Marinate clean slices of beef in a bowl with seasoning.
Step 4: Stir-fry the marinaded beef in a wok.
Step 5: Wash broccoli with water.
Step 6: Blanch clean broccoli in a cooking pot.
Step 7: Use a bowl to drain the blanched broccoli.
Step 8: Combine lightly cooked beef and blanched and drained broccoli in a wok.
Step 9: Prepare the seasoning: ginger, garlic, cooking oil, salt, light soy sauce, cooking wine, white pepper for the step: "Stir-fry the beef and broccoli mixture with the seasoning in a wok".
Step 10: Stir-fry the beef and broccoli mixture with the seasoning in a wok.
Step 11: Then, we get the cooked broccoli beef.

Carefully cooking at each step!
\end{lstlisting}

Risk management result from GPT-4:
\begin{lstlisting}[language=HTML]
Risk Management Plan:

Objective: To ensure a beneficial acquisition of Blizzard Entertainment by Microsoft while managing and mitigating potential risks.

I. Risk Identification

1. Financial Risks: A complete and thorough financial audit of Blizzard must be performed to determine potential financial risks, such bright as debts, pending lawsuit settlements, past or ongoing tax issues, declining revenue, etc.

2. Operational Risks: Understand Blizzard's operational weaknesses and strengths, and the impact of the acquisition on Microsoft's operations. 

3. Legal Risks: Review Blizzard's legal compliance status in all jurisdictions it operates in. Especially, risks related to intellectual property rights, privacy and data security laws, and labor laws must be evaluated.

4. Regulatory Risks: Determine any potential regulatory approvals Microsoft might need for the acquisition. 

II. Risk Assessment:

1. Quantify each identified risk wherever possible, in terms of potential financial impact, time delay, brand damage etc.
2. Determine the likelihood of each risk occurrence.
3. Prioritize the risks based on their potential impact and the likelihood of occurrence.

III. Risk Mitigation Planning:

1. Financial Risk: This could be mitigated through leverage and insurance, renegotiation of terms, or such other financial instruments.

2. Operational Risk: To mitigate this risk, Microsoft can plan to gradually integrate Blizzard, keeping key management personnel, or finding synergies with Microsoft's current operations.

3. Legal Risk: Microsoft should get warranties and indemnities from Blizzard's owners to cover potential post-acquisition legal disputes.

4. Regulatory Risk: Early engagement with regulators and professional advice can help smooth the acquisition process.

IV. Risk Monitoring: 

A system to continuously monitor and manage these risks should be set up. This could involve regular reporting, monitoring industry developments, legal changes etc.

V. Contingency Planning:

Establish contingency plans for high-impact risks. For instance, a backup plan if the acquisition fails, or if post-acquisition integration does not go as planned.

VI. Communication:

Continual communication with stakeholders to obtain feedback and to convey risk management efforts. This will ensure that all parties are aware of the risks and the measures being taken to mitigate them.

VII. Insurance Cover:

Consider adequate insurance coverage that includes Directors' and Officers' Liability and Professional Indemnity Insurance for all potential issues that might arise in connection with the proposed transaction.
   
The above steps will form the roadmap for Microsoft while negotiating terms for the acquisition of Blizzard Entertainment. Understanding risks before they come up ensures that Microsoft isn't surprised during negotiations, hence effectively mitigating them.
\end{lstlisting}

Risk management result after applying Formal-LLM:
\begin{lstlisting}[language=HTML]
Here is the final plan for Microsoft:

1: We need more information to ensure Shareholders of Blizzard Entertainment is entitled to sell Blizzard Entertainment. But we assume we have found out enough information to proceed the risk assessment process.
2: Then, based on current information, we believe Blizzard Entertainment is in good standing without material outstanding liabilities.
3: Then, based on current information, we believe neither CFIUS filings nor foreign investment filings in China are needed.
4: Then, we need to submit the regional antitrust filing of the United States and/or China. But we assume the filings will be approved to proceed the risk assessment process.
5: Pass the preliminary risk analysis. Details to be discussed in the transaction documents.

Note: This is a risk assessment process provided based on current information. Please ensure the accuracy of the provided information and possible additional supplemental information.
\end{lstlisting}

\section{Potential Risk}

Our research focuses on making LLM-based agent planning more controllable. To the best of our knowledge, we do not notice the potential risk of our research.

\section{Additional information of Datasets}

In our research, we used two benchmark datasets, \textbf{OpenAGI} \citep{openagi} and \textbf{TravelPlanner} \citep{xie2024travelplanner}. They are both under MIT license, which means these two datasets are publicly available. 

\textbf{OpenAGI} \citep{openagi} is used as the agent creation package to build agents for LLM-based agents. \textbf{TravelPlanner} \citep{xie2024travelplanner} is a benchmark crafted for evaluating language agents in tool-use and complex planning within multiple constraints. Our work is to explore the controllable planning ability of LLM-based agents, which is consistent with their intended use.

These two benchmark datasets are synthesized by algorithm and thus, do not contain any individual personal information.

The language of these two benchmark datasets is English.

For the number of examples and details of train / test / dev splits, we use the same dataset and follow the the same data splitting method with these two papers \citep{openagi, xie2024travelplanner}. 

\end{document}